%% file: main.tex
\documentclass[10pt,twocolumn,letterpaper]{article}

\usepackage[pagenumbers]{cvpr} 

\usepackage{multirow}
\usepackage{tabularx}

\definecolor{cvprblue}{rgb}{0.21,0.49,0.74}
\usepackage[pagebackref,breaklinks,colorlinks,allcolors=cvprblue]{hyperref}

\title{ WeatherCity: Urban Scene Reconstruction with Controllable \\
Multi-Weather Transformation}

\author{
Wenhua Wu$^*$ \quad  Huai Guan$^*$ \quad Zhe Liu \quad Hesheng Wang$^\dag$ \\ School of Automation and Intelligent Sensing, Shanghai Jiao Tong University, Shanghai, China
}

\begin{document}

\twocolumn[{%
\renewcommand\twocolumn[1][]{#1}%
    \setlength{\tabcolsep}{0.0mm} 
    \maketitle
    \begin{center}
\newcommand{\teaserwidth}{\textwidth}
    \vspace{-0.3in}
\includegraphics[width=\linewidth]{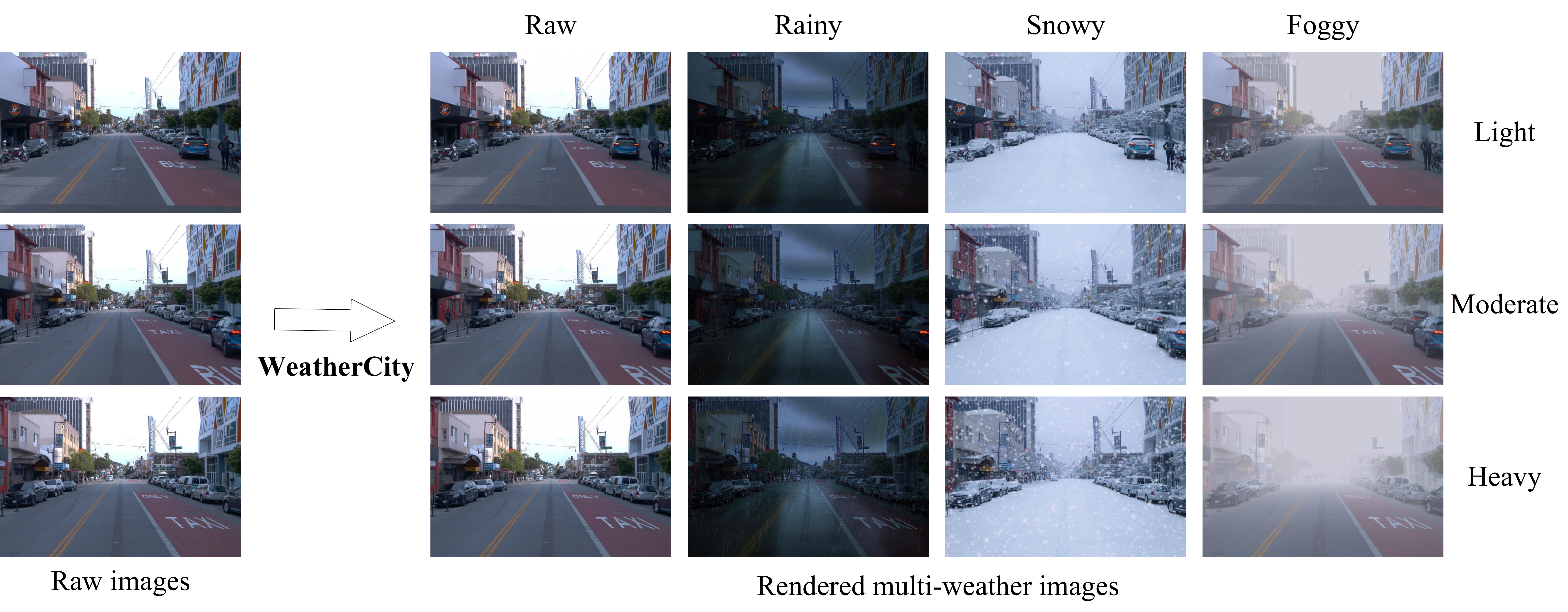}
      \vspace{-0.2in}
        \captionof{figure}{ 
        We present \textbf{WeatherCity}, a novel framework for dynamic urban scene reconstruction and controllable weather editing. Given a sequence of raw images, WeatherCity seamlessly integrates 4D reconstruction with flexible weather manipulation, producing highly consistent, photorealistic, and versatile multi-weather rendering results.
        }
    \label{fig:teaser}
    \end{center}
}]

\renewcommand{\thefootnote}{} 
\footnotetext{$^*$ Equal contribution.}
\footnotetext{
$^\dag$ Corresponding author.
}

\input{sec/0_abstract}    
\input{sec/1_intro}
\input{sec/2_related_work}

\input{sec/3_method}

\input{sec/4_experiment}
\input{sec/5_conclusion}


\section*{Acknowledgements}

This work was supported in part by National Key R\&D Program of China (Grant No.2024YFB4708900), the Natural Science Foundation of China under Grant 62225309, U24A20278, and 62361166632, Fundamental and Interdisciplinary Disciplines Breakthrough Plan of the Ministry of Education of China JYB2025XDXM117.

{
    \small
    \bibliographystyle{ieeenat_fullname}
    \bibliography{main}
}

\input{sec/X_suppl}
\end{document}

%% file: sec/0_abstract.tex
\begin{abstract}

Editable high-fidelity 4D scenes are crucial for autonomous driving, as they can be applied to end-to-end training and closed-loop simulation. However, existing reconstruction methods are primarily limited to replicating observed scenes and lack the capability for diverse weather simulation. While image-level weather editing methods tend to introduce scene artifacts and offer poor controllability over the weather effects. To address these limitations, we propose \textbf{WeatherCity}, a novel framework for 4D urban scene reconstruction and weather editing. Specifically, we leverage a text-guided image editing model to achieve flexible editing of image weather backgrounds. To tackle the challenge of multi-weather modeling, we introduce a novel weather Gaussian representation based on shared scene features and dedicated weather-specific decoders. This representation is further enhanced with a content consistency optimization, ensuring coherent modeling across different weather conditions. Additionally, we design a physics-driven model that simulates dynamic weather effects through particles and motion patterns.
Extensive experiments on multiple datasets and various scenes demonstrate that WeatherCity achieves flexible controllability, high fidelity, and temporal consistency in 4D reconstruction and weather editing. Our framework not only enables fine-grained control over weather conditions (e.g., light rain and heavy snow) but also supports object-level manipulation within the scene. Codes are released at \url{https://github.com/IRMVLab/WeatherCity}.

\end{abstract}

%% file: sec/1_intro.tex
\section{Introduction}
\label{sec:intro}

High-fidelity 4D simulation is crucial for autonomous driving, as it can not only provide diverse training samples covering edge cases like extreme weather but also construct reproducible virtual testing environments for closed-loop evaluation~\cite{christodoulides2025survey}. However, significant challenges persist across the "reconstruction-editing-simulation" pipeline for dynamic scenes, severely limiting the robustness of autonomous systems in complex environments.

First, existing 4D scene reconstruction methods struggle to overcome the limitation of observation dependency. While emerging techniques like Neural Radiance Fields (NeRF)~\cite{mildenhall2021nerf} and 3D Gaussian Splatting (3DGS)~\cite{kerbl20233d} achieve high-fidelity geometric and photometric reconstruction, they can only reproduce the weather conditions present during data capture, failing to simulate challenging scenarios such as rain, snow, or fog. Subsequent optimizations for urban scenes, such as StreetGaussians~\cite{yan2024street}, which models vehicle motion through dynamic appearance models, and OmniRe~\cite{chenomnire}, which introduces a multi-node structure to support modeling of pedestrians and deformable objects, primarily focus on "object-level adjustments" (e.g., modifying vehicle positions or counts) without addressing the crucial environmental dimension of weather.

Second, image-level weather editing techniques fail to meet the consistency requirements essential for 4D scene generation. Early approaches, primarily based on GAN architectures~\cite{goodfellow2020generative}, required training specialized models for each weather condition, resulting in limited editing diversity and poor generalization~\cite{zheng2024tpsence}. Recent methods have turned to diffusion models based~\cite{greenberg2023s2st, zhang2023adding}, which allow flexible weather control via text prompts. However, they often introduce content hallucinations, such as altering the position of road markings or distorting buildings. 
Furthermore, they offer limited capability for fine-grained control over weather intensity parameters, such as the strength of rain or snow or the density of fog. Recent methods have attempted weather editing based on 3D representation. DerainNeRF~\cite{li2024derainnerf} and WeatherGS~\cite{qian2025weathergs} utilize NeRF and 3D Gaussians to reconstruct and edit rainy scenes, but their functionality is limited to removing raindrops. ClimateNeRF~\cite{li2023climatenerf} can simulate static weather effects such as snow cover and flooding, but it cannot generate dynamic weather phenomena such as falling rain or snow. WeatherEdit~\cite{qian2026weatheredit} introduces a two-stage weather editing framework utilizing 4D Gaussians to simulate dynamic weather effects. However, its 3D reconstruction for different weather scenes is independent, without considering the consistency of the underlying scene geometry and content.

To address these challenges, we present WeatherCity, an editable high-fidelity 4D urban scene reconstruction and weather editing method. To achieve flexible scene editing, we employ a text-guided image editing model for weather image synthesis, which enables flexible generation of target weather conditions.
To ensure scene consistency, we propose a Weather Gaussian representation with shared features and multi-weather decoders that disentangle intrinsic structural and textural features of the scene from weather specific appearance attributes. This allows consistent scene structure across varying weather conditions while effectively modeling diverse weather effects. Furthermore, we introduce a content consistency loss to further enhance structural coherence. For simulating and controlling dynamic weather effects, we design a physics-driven dynamic weather simulation system. For rain and snow, we develop a variety of weather particles along with corresponding motion models. For fog, we implement depth-aware fog rendering based on the Beer–Lambert law. Our method achieves realistic dynamic weather simulation with fine-grained control over weather intensity, such as the amount of rain or snow, and the density of fog. 

The main contributions are summarized as follows:
\begin{itemize}
\item  We propose a unified framework supporting integrated 4D Reconstruction - Weather Editing - Dynamic Simulation, effectively elevating 2D image editing to 4D simulation and enabling the generation of multi-weather, highly consistent 4D scenes for autonomous driving applications.

\item We introduce weather Gaussian representation with shared feature and multi-weather decoders, which disentangles scene geometry from weather-related appearance. This ensures structural consistency across different weather conditions and facilitates efficient switching and editing of multi-weather scenes.

\item We construct a physics-driven dynamic weather simulation system, designing weather effects based on weather particles and optical principles for rain, snow, and fog respectively, thereby achieving dynamic weather simulation with both visual realism and physical consistency while enabling precise control.
 \end{itemize}

%% file: sec/2_related_work.tex
\section{Related Work}
\label{sec:related_work}

\subsection{Urban Scene Reconstruction}

The field of 3D scene reconstruction has evolved significantly from traditional geometric methods~\cite{ozyecsil2017survey, wang2021multi} to modern neural representations~\cite{mildenhall2021nerf, kerbl20233d, ost2021neural, zhou2024drivinggaussian, yan2024street, chenomnire, wu2024emie, wu2025bev,chen2023periodic,cui2025streetsurfgs, deng2025best, deng2025vpgs, zhu20243d,su2025dial}. 
The introduction of NeRF~\cite{mildenhall2021nerf} and 3DGS~\cite{kerbl20233d} marked a significant leap forward, enabling dense and photorealistic reconstruction of complex driving environments. Neural Scene Graphs (NSG)~\cite{ost2021neural} proposed a graph-based representation that separately models static backgrounds and dynamic objects using dedicated radiance fields. EmerNeRF~\cite{yang2023emernerf} further advanced this line by introducing a self-supervised framework that automatically decomposes scenes into static and dynamic components. 
DrivingGaussian~\cite{zhou2024drivinggaussian} developed an incremental static Gaussian representation combined with dynamic Gaussian graphs to enable object-level scene editing, while Street Gaussians~\cite{yan2024street} achieved more accurate dynamic reconstruction through object pose optimization and 4D spherical harmonics. OmniRe~\cite{chenomnire} further pushed the boundaries by introducing deformable nodes to handle non-rigid dynamic objects like pedestrians. Despite these advances, these methods inherently bake in the weather conditions present during data capture, lacking the ability to disentangle and control meteorological factors. 

\subsection{Image-level Weather Editing}
Weather editing and enhancement in 2D images represent a standing challenge in computer vision. With the development of deep learning, significant progress has been made~\cite{zheng2024tpsence,zhang2023adding,deutch2024turboedit, sheynin2024emu, brooks2023instructpix2pix,geyer2023tokenflow,esser2024scaling}. ClimateGAN~\cite{schmidtclimategan} introduced a method for realistic flood simulation on real-world images. WeatherGAN~\cite{li2022weather} built upon StarGAN v2~\cite{choi2020stargan} and proposed a weather feature-guided approach for multi-domain translation. TPSeNCE~\cite{zheng2024tpsence} further improved weather editing consistency by introducing a triangular probability similarity constraint. The emergence of Diffusion Models has substantially advanced image generation capabilities, enabling more flexible and controllable editing. For instance, ControlNet~\cite{zhang2023adding} incorporated spatial conditioning into text-to-image diffusion models, facilitating various conditional image edits. InstructPix2Pix~\cite{brooks2023instructpix2pix} combined language instructions with diffusion models to enable instruction-based image editing, leveraging large-scale generated data for training. TurboEdit~\cite{deutch2024turboedit} reduced editing artifacts via shifted noise scheduling and novel guidance techniques. Despite these improvements in image quality, challenges such as artifacts and inconsistencies persist. Moreover, these methods generally lack fine-grained control over weather intensity and effects.

\subsection{3D-level Weather Editing}
Recent research has begun to explore weather editing directly in 3D scene representations~\cite{qian2026weatheredit}. ClimateNeRF~\cite{li2023climatenerf} integrates physical simulation with NeRF to achieve editing of various climate effects, although it is limited to static weather phenomena and cannot simulate dynamic conditions. WeatherGS~\cite{qian2025weathergs} focuses on mitigating weather artifacts, restoring clear scenes from adverse weather inputs. StyleGaussian~\cite{liu2024stylegaussian} and SGSST~\cite{galerne2025sgsst} adapt 3D Gaussians for artistic style transfer using reference images, yet their frameworks are not tailored for realistic weather editing. RainyGS~\cite{dai2025rainygs} combines physical simulation with Gaussian splatting to realistically simulate rainy conditions. Similarly, both Let it Snow~\cite{fiebelman2025let} and Weather-Magician~\cite{sang2025weather} utilize 3D Gaussians to simulate dynamic weather effects, but they primarily focus on foreground weather particles and lack synchronized background weather editing, limiting the overall realism. WeatherEdit~\cite{qian2026weatheredit} introduces a two-stage framework that considers background weather and dynamic weather particle editing. However, its 3D reconstruction for different weather scenes is independent.
In contrast, our method considers the consistency of scene geometry and content across different weather conditions.

%% file: sec/3_method.tex
\section{Method}\label{sec:method}

\begin{figure*}[t] 
\center
\includegraphics[width=1.0\textwidth]{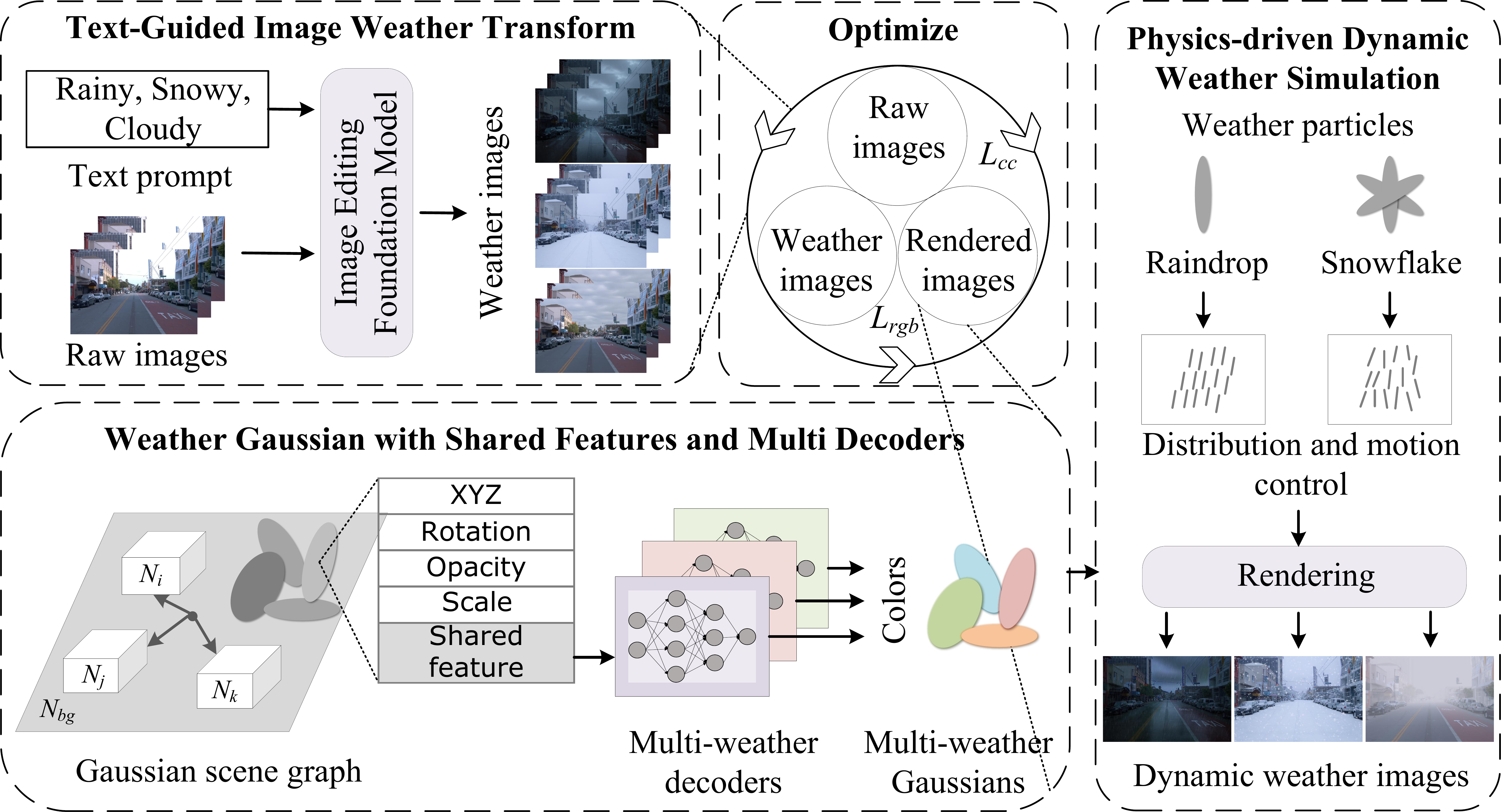}
\caption{Overview of WeatherCity. Our framework comprises four main modules. First, the image editing module employs a text-guided video editing foundation model to adapt image weather background. Second, the scene representation module introduces a weather-aware Gaussian representation based on shared features and multi-weather decoders, which disentangles geometric-textural attributes from weather-specific appearances, thereby ensuring structural consistency across varying meteorological conditions. Subsequently, we construct RGB and content losses for consistency optimization. Finally, a physics-driven dynamic weather simulation mechanism is designed to achieve flexible and controllable editing of diverse dynamic weather effects.
}
\label{framework}
\end{figure*}

Given a sequence of captured raw scene data, WeatherCity achieves joint 4D dynamic reconstruction and flexible, controllable weather editing. Our framework, illustrated in Fig.~\ref{framework}, comprises four core components. Firstly, the image editing module leverages a text-guided image editing foundation model to flexibly alter image weather backgrounds while preserving the original scene content~\ref{sec_3_1}. Subsequently, the scene representation module introduces a novel Weather Gaussian Representation, which builds upon shared scene features and weather-specific decoders to disentangle geometric and textural attributes from weather appearances, thereby ensuring structural consistency across different conditions~\ref{sec_3_2}. Consistency is further optimized by employing RGB and content losses between the rendered and edited multi-weather images~\ref{sec_3_3}. Finally, a physics-driven dynamic weather simulation module is designed, which utilizes particle systems to simulate rain and snow and applies the Beer-Lambert law to model fog, enabling controllable editing of various dynamic weather effects~\ref{sec_3_4}. 

\subsection{Text-Guided Image Weather Background Editing}
\label{sec_3_1}
The first stage of our pipeline involves editing the input images $\{I_t^{raw}| I_t^{raw} \in \mathbb{R}^{ H\times W \times 3}, t = 1,..., N\}$ to exhibit various target weather conditions. These edited sequences serve as crucial supervision signals for the subsequent 3D reconstruction and editing tasks. This requires the edited images to achieve both realistic weather effects and consistency with the original scene content.

To this end, we leverage Qwen-Image~\cite{wu2025qwen}, a powerful text-guided image editing model. For each desired weather condition, we design a corresponding text prompt. Our prompts are meticulously crafted to not only describe the target weather effect (e.g., "a rainy city street") but also to explicitly emphasize the strict preservation of the original scene content. Benefiting from these carefully designed prompts and the robust image editing capabilities of Qwen-Image~\cite{wu2025qwen}, we obtain highly realistic and temporally consistent multi-weather image sequences $\{I_t^w| w \in \mathcal{W},t = 1,..., N\}$, where $\mathcal{W}$ is the set of multi weathers.

\subsection{Weather Gaussian with Shared Feature and Multi-Weather Decoders}
\label{sec_3_2}
Following OmniRe~\cite{chenomnire}, we employ a dynamic Gaussian graph to structure the scene, enabling flexible modeling and control of movable objects. Our scene graph $\mathcal{G} = \{\mathcal{N}, \mathcal{E}\}$ comprises the following nodes $\mathcal{N}$:

\begin{itemize}
    \item Sky Node $\mathcal{N}_{sky}$: Representing the distant sky via an optimizable environment texture map.
    \item Background Node $\mathcal{N}_{bg}$: Modeling the static background composed of 3D Gaussians.
    \item Rigid Nodes $\mathcal{N}_{rigid}$: Representing movable rigid objects (primarily vehicles), each composed of 3D Gaussians.
    \item Non-Rigid Nodes $\mathcal{N}_{nonrigid}$: Accounting for deformable objects (primarily pedestrians), each composed of deformation 3D Gaussians.
\end{itemize}

To achieve consistent scene reconstruction and diverse weather editing, we design a novel Weather Gaussian Representation that disentangles inherent scene geometry from weather-dependent appearance.

Each Gaussian primitive $G_i$ in nodes $\{\mathcal{N}_{bg}, \mathcal{N}_{rigid}, \mathcal{N}_{nonrigid}\}$ is parameterized by fellows:
\begin{equation}
    G_i = \{\mu_i, s_i, r_i, o_i, f_i\},
\end{equation}
where:
\begin{itemize}
    \item $\mu_i \in \mathbb{R}^3$ denotes the 3D center position.
    \item $s_i \in \mathbb{R}^3$ represents the scale factors along three axes.
    \item $r_i \in \mathbb{R}^4$ is the rotation quaternion.
    \item $o_i \in [0, 1]$ is the opacity value.
    \item $f_i \in \mathbb{R}^d$ is a shared appearance feature encoding the intrinsic texture and material properties.
\end{itemize}

For each weather condition $w \in \mathcal{W}$, the corresponding Gaussian color $c_i^w$ is decoded by a weather-specific MLP $\phi_w$:
\begin{equation}
    c_i^w = \phi_w(f_i).
\end{equation}

\begin{figure*}[t] 
\center
\includegraphics[width=1.0\textwidth]{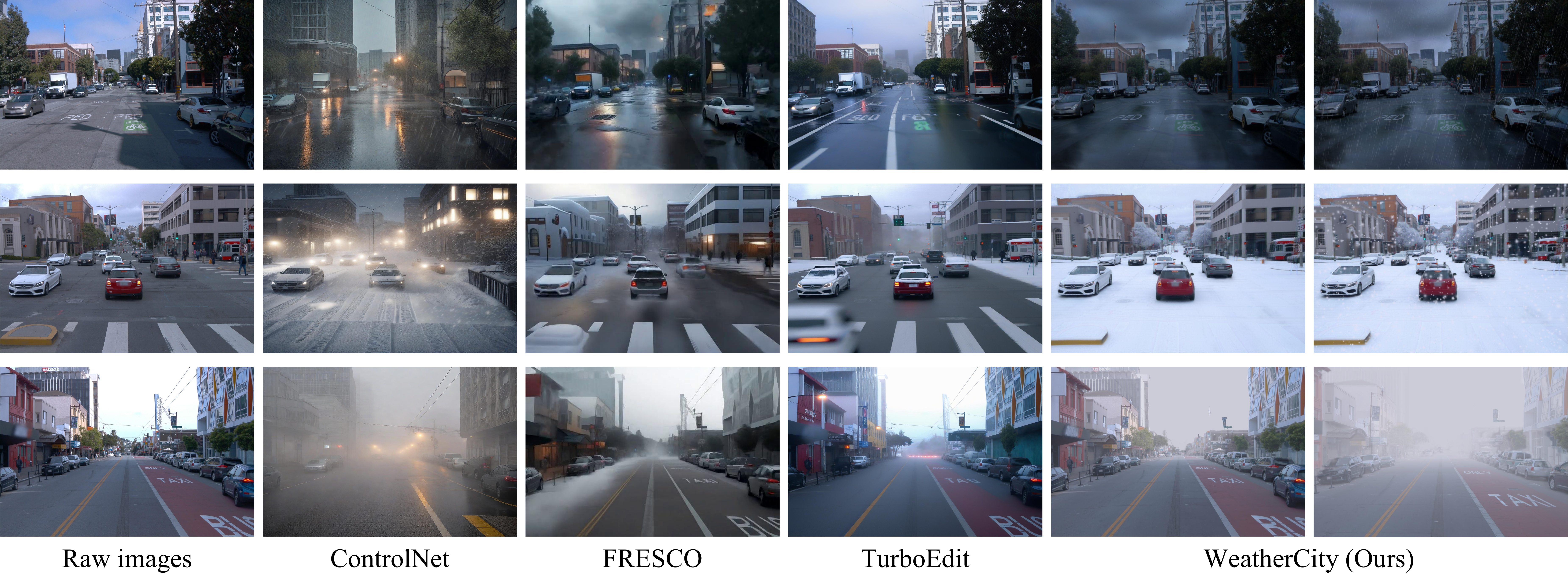}
\caption{Qualitative results of multi-weather editing on the Waymo Open Dataset. Our method produces realistic and consistent weather editing effects while supporting dynamic weather simulation, whereas all baselines exhibit significant scene distortion.}
\label{fig_wamo}
\end{figure*}

Subsequently, the shared feature Gaussians are transformed into multi-weather Gaussians $G_i^w = \{\mu_i,\Sigma_i,o_i,c_i^w\}$ through our dedicated weather-specific decoders. The covariance matrix $\Sigma \in \mathbb{R}^{3 \times 3} = RSS^\top R^\top$. The spatial influence at point $x$ given by:
\begin{equation}
    g_i(x) = e^{-\frac{1}{2}(x - \mu_i)^\top \Sigma^{-1}(x - \mu_i)}.
\end{equation}

During rendering, each 3D Gaussian is first projected onto the camera coordinate. The corresponding 2D covariance matrix $\boldsymbol{\Sigma}'$ in the image plane is derived as:
\begin{equation}
    \Sigma' = J W \Sigma W^\top J^\top
\end{equation}
where $J$ denotes the Jacobian of the projective transformation and $W$ represents the view transformation matrix. For each pixel, the contributing Gaussians are sorted by depth and rendered via alpha blending:
\begin{equation}
    \hat{I}_t^w = \sum_{i } c_i^w \alpha_i \prod_{j=1}^{i-1} (1 - \alpha_j), \quad     \hat{D}_t = \sum_{i } d_i \alpha_i \prod_{j=1}^{i-1} (1 - \alpha_j),
    \label{rendering}
\end{equation}
where $\alpha_i$ is the computed opacity of the $i$-th Gaussian after sorting, $d_i$ is the depth of the depth of the $i$-th Gaussian.

This design forces the shared features to capture the intrinsic textural properties of the scene, while the separate decoders learn to model the photometric impact of specific weather conditions. Consequently, our representation ensures structural consistency across different weathers on one hand, and enables distinct environmental photometric modeling on the other.

\subsection{Multi-Weather Consistency Optimization}
\label{sec_3_3}
We jointly optimize our Weather Gaussian Representation using a composite loss function that aligns the rendered scenes with both the original and the edited multi-weather images.

\noindent\textbf{RGB Loss}. We render the scene under both the original clear weather and edited multi- weather $\mathcal{W}$, and compute the RGB loss against the corresponding ground truth and edited images. The RGB loss is defined as:
\begin{equation}
\begin{aligned}
        \mathcal{L}_{rgb} = \sum_{t=1}^N \sum_{w \in \mathcal{W} \cup \{raw\}} &(1-\lambda) \left\| \hat{I}_{t}^w - I_{t}^w \right\|_1 \\
        & +\lambda(1-{\rm SSIM}(\hat{I}_{t}^w,I_{t}^w)).
\end{aligned}
\end{equation}

\noindent\textbf{Content Consistency Loss}. To further enforce semantic and structural consistency across different weather conditions, we introduce a content consistency loss. Specifically, we employ a pre-trained VGG network~\cite{Simonyan2014VeryDC} $\Phi$ to extract content features from the rendered images and the original weather images. The content consistency loss is computed as the L1 distance between the feature of the rendered image under weather condition $w$ and the original weather image:
\begin{equation}
\mathcal{L}_{cc} = \sum_{t=1}^N\sum_{w \in\mathcal{W}}\|\Phi(\hat{I}_t^w) - \Phi(I_t^{raw})\|.
\end{equation}

Applying 2D image editing models on a frame-by-frame may leads to temporal flickering and geometric inconsistencies.
This loss is able to optimize regions with scene content distortion in the 2D editing results, preventing inconsistent 2D edits from compromising scene coherence. This ensures that modifying the weather attributes $w$ does not alter the underlying scene content.

\begin{figure*}[t] 
\center
\includegraphics[width=1.0\textwidth]{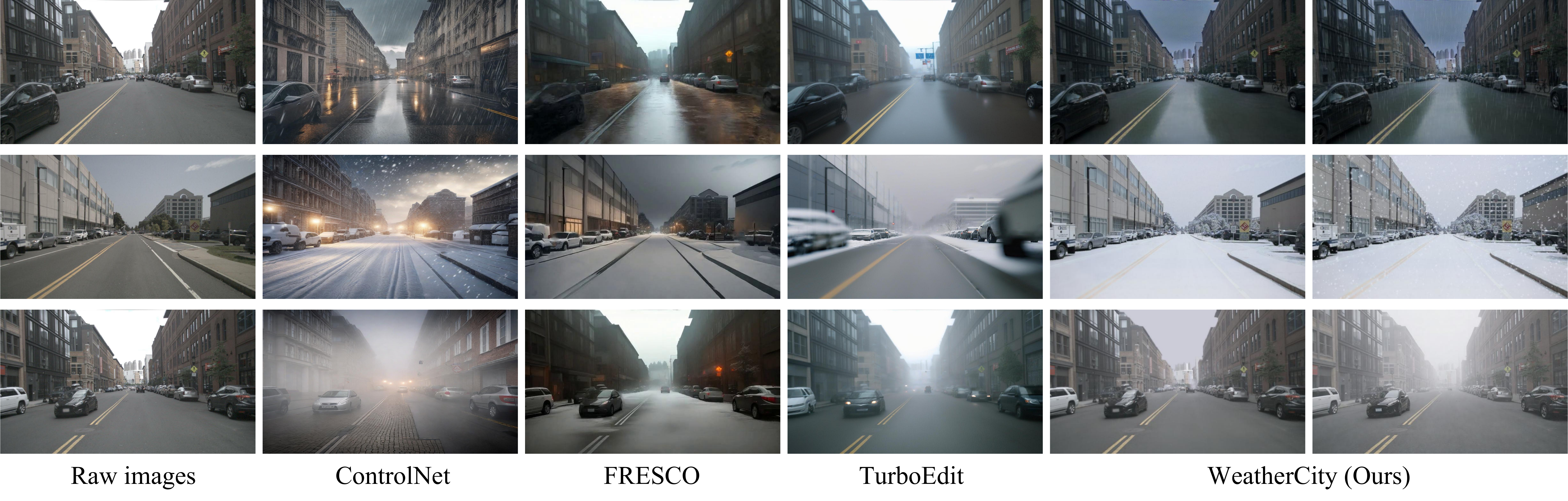}
\caption{Qualitative results of multi-weather editing on the nuScenes dataset. Our method produces realistic and consistent weather editing effects while supporting dynamic weather simulation, whereas all baselines exhibit significant scene distortion.}
\label{fig_nuscene}
\end{figure*}

\noindent\textbf{Depth Loss:} To supervise the geometric information of the scene, we compute an L1 loss between the rendered depth and the sparse depth map $D_t$ obtained from LiDAR projection:
\begin{equation}
\mathcal{L}_{depth} = \sum_{t=1}^N\|\hat{D}_{t} - D_{t}\|.
\end{equation}

 The complete optimization loss is a weighted combination of all loss terms:
\begin{equation}
\begin{aligned}
\mathcal{L}_{total} = & \mathcal{L}_{rgb} + \lambda_{cc}\mathcal{L}_{cc} + \lambda_{depth}\mathcal{L}_{depth} \\
 & + \lambda_{opacity}\mathcal{L}_{opacity}+\mathcal{L}_{reg},
\end{aligned}
\end{equation}
where $\lambda_{cc}$, $\lambda_{depth}$, and $\lambda_{opacity}$ are balancing weights for the respective loss components. $\mathcal{L}_{opacity}$ is the opacity loss, which ensures the Gaussian opacities align with the non-sky mask, and $\mathcal{L}_{reg}$ is the regularization loss. More details are available in the Supplementary Material.

\subsection{Physics-Driven Dynamic Weather Simulation}
\label{sec_3_4}
To achieve visually realistic, physically consistent, and finely controllable dynamic weather simulation, we develop a physics-driven dynamic weather simulation system. 


\noindent \textbf{Weather Particle Modeling.} We model rain and snow particles using Gaussian ellipsoids: raindrops are represented by a single elongated Gaussian to capture their vertically stretched characteristics and motion blur effects, while snowflakes are approximated by three concentric Gaussian ellipsoids with identical scales arranged at 60-degree angles to form a basic crystal shape. 
The scale parameters of the Gaussians control the size of the weather particles, while the rotation parameters govern their orientation. Furthermore, the spatial distribution of weather particles is key to controlling weather intensity. To this end, we construct a spatial bounding volume based on the reconstructed scene. Within this volume, particle attributes, including position $\mu_i$, rotation $r_i$, and opacity $o_i$, are initialized according to varying densities to achieve natural visual variations.

\noindent \textbf{Motion Control.}
Each particle moves according to a velocity vector $\mathbf{v}$ updated per frame based on physics-driven parameters. 
The velocity equations for raindrop and snowflake particles are given as follows:
\begin{equation}
    \mathbf{v}_\text{rain} = \mathbf{v}_\text{fall} + \mathbf{v}_\text{wind}, \quad\mathbf{v}_\text{snow} = \mathbf{v}_\text{fall} + \mathbf{v}_\text{wind} + \mathbf{v}_\text{turb},
\end{equation}
where $\mathbf{v}_\text{fall}$ is a constant downward velocity, $\mathbf{v}_\text{wind}$ is a global wind vector parameterized by magnitude $v_\text{mag}$, tilt angle $\theta_\text{wind}$, and azimuth $\phi_\text{wind}$, the elongated Gaussian is dynamically aligned with $\mathbf{v}$ to reflect wind-influenced trajectories. $\mathbf{v}_\text{turb}$ is a stochastic turbulence component applied only to snowflakes, which induces their characteristic fluttering and non-linear descent.

\noindent \textbf{Unified Weather Rendering.}
A core advantage of our method lies in its unified rendering pipeline. Instead of being rendered through a separate pass, the dynamically generated weather Gaussians for rain and snow are directly integrated into the dynamic Gaussian scene graph as weather nodes $\mathcal{N}_{rain}$ and $\mathcal{N}_{snow}$, thereby facilitating subsequent scene editing and manipulation.
All Gaussians (e.g, scene and weather particles) are rasterized together using the standard Gaussian Splatting process. This unified formulation naturally ensures correct occlusion, composition, and blending, resulting in seamless and physically consistent integration of weather effects.

Since fog is uniformly diffused throughout the space, we implement depth-aware fog simulation based on the Beer-Lambert law~\cite{swinehart1962beer}. The final fog-affected color $c_{render}^{fog}$ is obtained by blending the rendered pixel color $c_{render}$ with the global fog color $c_{fog}$ :
\begin{equation}
    c_{render}^{fog} = f c_{render} +(1-f)c_{fog},
\end{equation}
where $f = e^{-d_f d_{render}}$ denotes the transmittance, $d_f$ is the fog density parameter, and $d_{render}$ represents the depth value obtained from Equation~\ref{rendering}. By adjusting parameters $c_{fog}$ and $d_f$, we achieve realistic rendering of fog effects with varying density and colors.

%% file: sec/4_experiment.tex
\section{Experiment}
\label{sec:experiment}
We conduct extensive experiments to comprehensively evaluate the effectiveness and superiority of WeatherCity. 

\begin{table*}[t]
\caption{Quantitative comparison on Waymo Open Dataset and nuScenes Dataset. $\uparrow$ means higher is better. WeatherCity achieves state-of-the-art performance across all metrics, outperforming other methods by a large margin.
}
\label{tab_all_results}
\centering
\begin{tabular}{lcccccc}
\toprule
\multirow{2}{*}{\textbf{Method}} & \multicolumn{3}{c}{\textbf{Waymo Open Dataset}} &
\multicolumn{3}{c}{\textbf{nuScenes Dataset}} \\
\cmidrule(lr){2-4}\cmidrule(lr){5-7}
& CLIP-S$\uparrow$ & CLIP-DS$\uparrow$ & Sem-CS$\uparrow$ 
& CLIP-S$\uparrow$ & CLIP-DS$\uparrow$ & Sem-CS$\uparrow$ 
\\

\midrule

ControlNet~\cite{zhang2023adding} 
& 0.634 & 0.238 & 0.695 & 0.656 & 0.228 & 0.811  \\

TurboEdit~\cite{deutch2024turboedit} 
& 0.830 & 0.220 & 0.801 & 0.782 & 0.250 & 0.829 \\

FRESCO~\cite{yang2024fresco} 
& 0.720 & 0.213 & 0.824 & 0.710 & 0.224 & 0.855  \\

Qwen-Image~\cite{wu2025qwen} 
& 0.785 & 0.279 & 0.843 & 0.804 & 0.279 & 0.902  \\

\textbf{WeatherCity (Ours)} 
& \textbf{0.872} & \textbf{0.303} & \textbf{0.915} 
& \textbf{0.870} & \textbf{0.302} & \textbf{0.968}  \\

\bottomrule
\end{tabular}
\end{table*}

\subsection{Experimental Setting}
\label{sec:exp_set}
\noindent \textbf{Dataset.}
We evaluate our method on two prominent autonomous driving benchmarks: the Waymo Open Dataset~\cite{sun2020scalability} and the nuScenes dataset~\cite{caesar2020nuscenes}. Both datasets provide diverse driving scenarios with multi-sensor data including multi-view images and LiDAR point clouds. For quantitative evaluation, we select five representative scenes rich in dynamic objects, each comprising 30 consecutive frames. The image resolutions are $1920\times 1080$ for Waymo and $1600\times 900$ for nuScenes.

\noindent \textbf{Evaluation Metrics.}
To assess weather editing quality, we employ the following metrics:
\begin{itemize}
\item \textbf{CLIP-Score (CLIP-S):} Measures content preservation by computing the cosine similarity between CLIP image embeddings of edited and original images~\cite{hessel2021clipscore}.
\item \textbf{CLIP Directional Similarity (CLIP-DS):} Evaluates alignment between edited images and target text prompts in the CLIP embedding space~\cite{gal2022stylegan}.
\item \textbf{Semantic Consistency Score (Sem-CS):} Measures the semantic consistency between edited and original images using a frequency-weighted Intersection over Union computed from a semantic segmentation model~\cite{liu2022convnet}.
\end{itemize}

\noindent \textbf{Baselines.} 
To demonstrate the advantages of our method, we compare it against several state-of-the-art approaches, including image editing methods, ControlNet~\cite{zhang2023adding} and TurboEdit~\cite{deutch2024turboedit}, and the video editing model FRESCO~\cite{yang2024fresco}. For a fair comparison, all methods are conditioned on the same textual prompts. 

\noindent \textbf{Implementation Details.}
The shared Gaussian feature dimension is set to 32, and the weather-specific MLP decoder consists of two linear layers with ReLU activation and a Sigmoid output mapping features to RGB. We train the model using Adam for 30,000 iterations with a learning rate of $1\mathrm{e}{-4}$, using loss weights $\lambda_{cc}=1.0$, $\lambda_{depth}=0.01$, and $\lambda_{SSIM}=0.2$. The content loss uses VGG-19~\cite{Simonyan2014VeryDC} \texttt{relu\_4\_1} features. Dynamic weather effects use 40,000 particles for rain and 16,000 for snow. Fog color and density parameters are set to $c_{fog}=[0.80,\,0.80,\,0.85]$ and $d_f=0.2$. All Gaussians are rasterized together using the standard Gaussian Splatting pipeline. All experiments are conducted on a server with an Intel W-3335 CPU and an RTX 8000 GPU.
More details are provided in the supplementary material.

\subsection{Experimental Results}
\label{sec:exp_res}
\noindent \textbf{Multi-Weather Editing.}
Tab.~\ref{tab_all_results} summarizes the quantitative results for multi-weather editing on the Waymo and nuScenes datasets. Our method achieves state-of-the-art performance across all metrics, significantly outperforming all baseline approaches. The notable improvements in CLIP-S and Sem-CS metrics particularly demonstrate our method's superior capability in preserving scene content and semantic consistency. It is strongly supported by the qualitative results shown in Fig.\ref{fig_wamo} and Fig.~\ref{fig_nuscene}. Our method precisely maintains the scene's geometric structure and semantic content while generating highly realistic weather effects, including overcast skies, wet ground reflections, accumulated snow on surfaces, and depth-attenuated fog. Furthermore, our method generates convincing dynamic weather particles (raindrops and snowflakes) achieving a level of realism unattainable by image-level editing methods. Although baselines can produce certain weather appearances, they introduce severe content distortion—manifested as warped vehicles, hallucinated structures, and erroneous lane markings. Additionally, image-level editing methods fundamentally lack the capability to produce depth-aware atmospheric effects. 




\begin{figure}[t] 
\center
\includegraphics[width=1.0\linewidth]{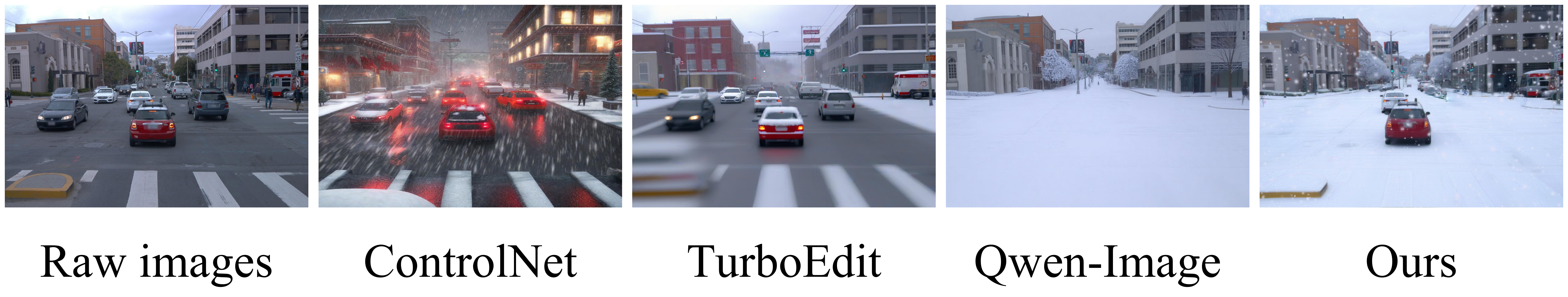}
\caption{Visualization of object editing results. The text prompt is "Remove all vehicles except the red and white ones in the center and change the weather to snowy". }
\label{fig_edit}
\end{figure}

\noindent \textbf{Object-Level Scene Editing}. Our framework supports not only weather editing but also object-level manipulation. Leveraging the dynamic Gaussian scene graph, we achieve precise control over dynamic nodes in the scene, enabling object removal, insertion, and repositioning. Fig~\ref{fig_edit} presents a visual comparison of object editing results. Using the text prompt: "Remove all vehicles except the red and white ones in the center and change the weather to snowy", our method accurately executes the requested edits. In contrast, baseline approaches either fail to remove the specified vehicles or eliminate them entirely without discrimination, demonstrating the superior precision of our editing capability.

\begin{table}[h] 
\caption{Comparison of runtime measured in FPS.
}
\label{tab:runtime}
\centering
\begin{tabular}{lc}
\toprule
Method & Speed (FPS) $\uparrow$ \\
\midrule
ControlNet~\cite{zhang2023adding} 
& 0.033 \\
TurboEdit~\cite{deutch2024turboedit} 
& 0.097 \\
FRESCO~\cite{yang2024fresco} 
& 0.142 \\

WeatherCity (Ours) 
& \textbf{25.67} \\
\bottomrule
\end{tabular}
\end{table}

\noindent \textbf{Runtime Analysis.}
We compare the runtime performance in Tab~\ref{tab:runtime}. For image and video editing models, we report the average inference speed (in FPS). For WeatherCity, We provide the average rendering speed. WeatherCity achieves a rendering speed of 25.67 FPS, which is sufficient to meet the requirements for real-time simulation. 

\subsection{Ablation Study}

\begin{figure}[t] 
\center
\includegraphics[width=1.0\linewidth]{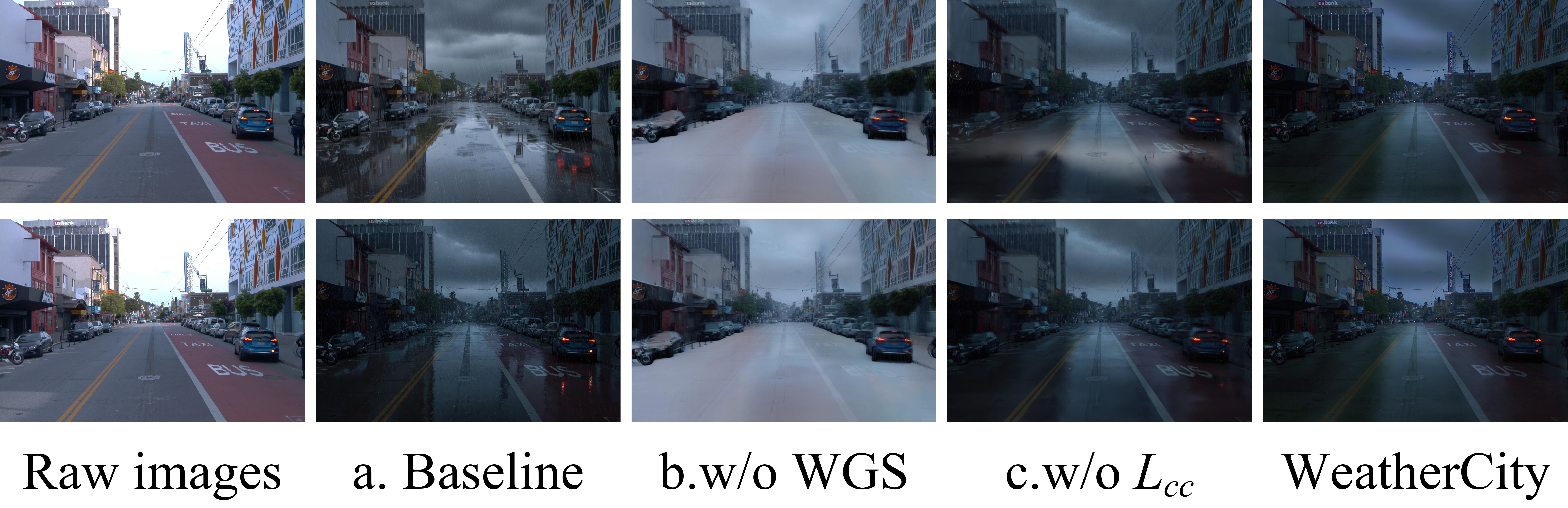}
\caption{Visualization of ablation study. }
\label{fig_ablation}
\end{figure}

To validate the effectiveness of each core component in WeatherCity, we conduct extensive ablation experiments under the following configurations:

\begin{enumerate}[label=\alph*.]
    \item Baseline: All proposed modules are removed, using only Qwen-Image for image editing.
    \item w/o WGS: Replace the weather Gaussian with the original 3D Gaussian, where all weather conditions share the same set of Gaussians.
    
    \item w/o $\mathcal{L}_{cc}$: The content consistency loss $\mathcal{L}_{cc}$ is removed.
\end{enumerate}
Tab.~\ref{tab_ablation} and Fig.~\ref{fig_ablation} present quantitative comparisons and visualizations across these configurations.

\begin{table}[h]
\caption{Ablation study results. The results validate the effectiveness of each design.
}
\label{tab_ablation}
\centering
\begin{tabular}{lccc}
\toprule
Method & CLIP-S $\uparrow$ &CLIP-DS $\uparrow$ & Sem-CS $\uparrow$ \\
\midrule
a. Baseline 
& 0.735 & 0.276 & 0.891 \\
b. w/o WGS 
& 0.781 & 0.212 & 0.894\\
c. w/o $\mathcal{L}_{cc}$ 
& 0.817 & 0.289 & 0.916 \\
WeatherCity & \textbf{0.880} & \textbf{0.320} & \textbf{0.943} \\
\bottomrule
\end{tabular}
\end{table}

 \noindent \textbf{Effectiveness of Weather Gaussian.} The removal of our Weather-aware Gaussian causes noticeable degradation in reconstruction metrics. The model fails to disentangle intrinsic scene textures from weather-specific appearances, leading to effect blending across different conditions as illustrated in Fig.~\ref{fig_ablation} (b). This confirms that our shared-scene-feature and weather-specific decoder design effectively separates inherent scene attributes from transient weather appearances, enabling stable scene structure preservation and distinct weather effect modeling.

 \noindent \textbf{Effectiveness of content consistency loss.} The removal of the content consistency loss leads to a notable decline in scene consistency metrics and introduces inconsistencies in the editing results, as illustrated in Fig.~\ref{fig_ablation} (c). These artifacts stem from frame-wise inconsistencies introduced by the Qwen-Image editing process as illustrated in Fig.~\ref{fig_ablation} (a). In contrast, the content consistency loss enforces feature alignment between the rendered images and the original scene through a pre-trained VGG network. This effectively rectifies the local artifacts caused by per-frame Qwen-Image editing, significantly enhancing semantic coherence and geometric integrity across weather transitions.

\begin{figure}[t] 
\center
\includegraphics[width=1.0\linewidth]{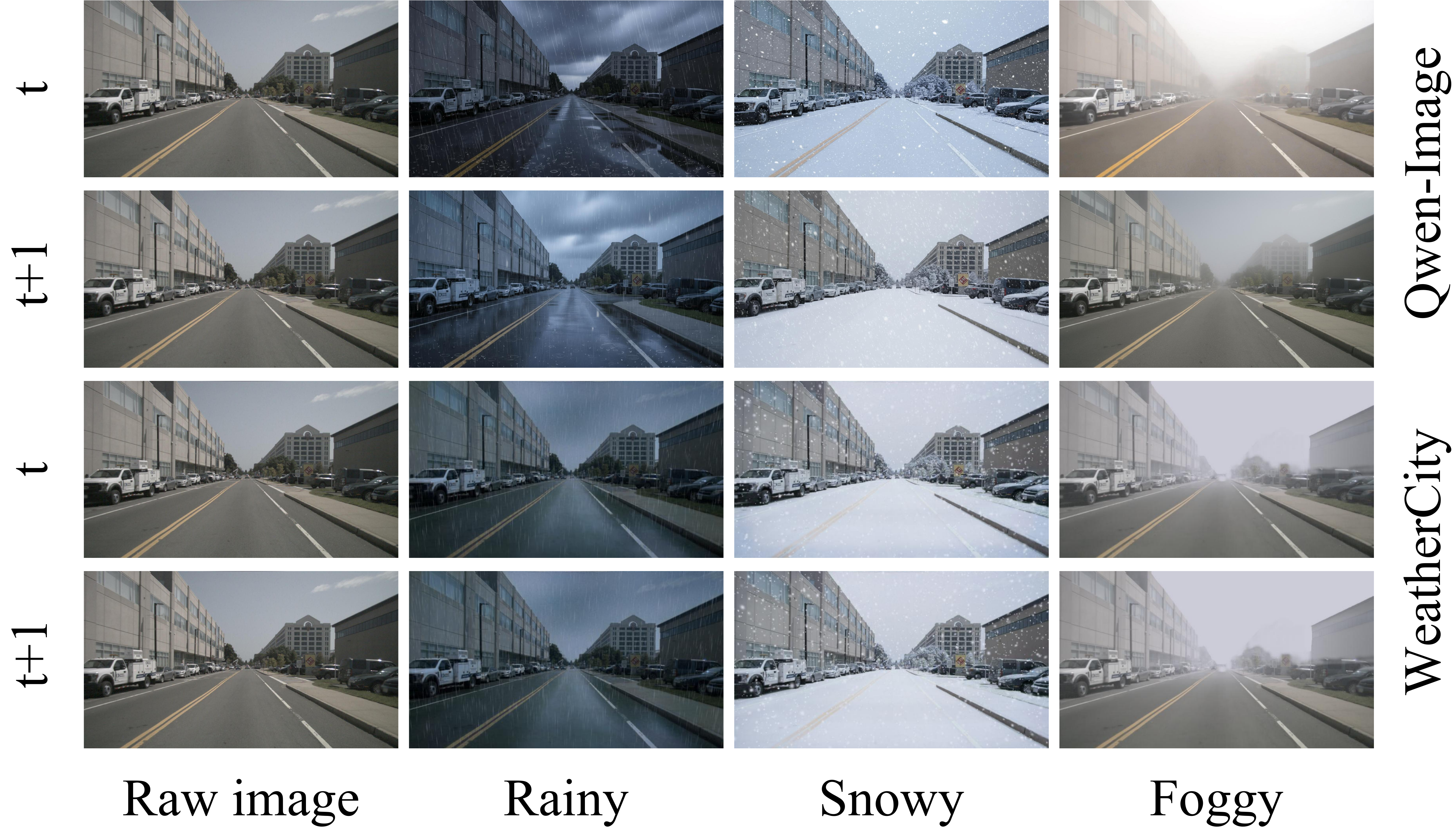}
\caption{Qualitative comparison on dynamic weather effect.}
\label{fig_particles}
\end{figure}

 \noindent \textbf{Effectiveness of physics-driven dynamic weather simulation.} To further demonstrate the advantages of our physics-driven dynamic weather module, we design additional comparative experiments using dynamic weather prompts (e.g., "heavy rain with raindrops falling under gravity, snowflakes drifting with weak wind, fog gradually thickening in the distance"). As shown in Fig.~\ref{fig_particles}, we compare the image editing results from Qwen-Image with our physics-based simulation. The visual comparison clearly reveals that the dynamic weather effects generated by Qwen-Image~\cite{wu2025qwen} lack temporal coherence, whereas our method achieves smooth inter-frame transitions through particle motion equations. 
 

%% file: sec/5_conclusion.tex
\section{Conclusion}
\label{sec:conclusion}

We present WeatherCity, a unified framework for high-fidelity 4D dynamic scene reconstruction and controllable weather simulation. We extend 2D image editing to 4D scene editing and propose a novel weather Gaussian representation that disentangles scene structure from weather appearance, and a physics-driven simulation system for dynamic effects. This enables the generation of diverse, temporally consistent, and photorealistic urban scenes under various weather conditions with fine-grained control. Extensive experiments validate that our method outperforms alternatives in visual quality, cross-weather consistency, and editing flexibility. WeatherCity not only provides a powerful tool for autonomous driving simulation but also establishes a solid foundation for future research in dynamic and controllable virtual environment creation. Limitations. The current system requires manual tuning of weather particle parameters. Future work will focus on developing more automated editing algorithms to streamline this process.

%% file: sec/X_suppl.tex
\clearpage
\setcounter{page}{1}
\maketitlesupplementary


\begin{abstract}
    In the supplementary material, we present additional implementation details (Sec.~\ref{sec_detail}), including training (Sec.~\ref{subsec_traing}), baselines (Sec.~\ref{subsec_base}) and evaluation (Sec.~\ref{subsec_eval}). We also provide further experimental results and analysis (Sec.~\ref{sec_res}), including detailed quantitative and qualitative results (Sec.~\ref{subsec_res}), temporal consistency comparison (Sec.~\ref{subsec_tem}), and 3D baseline comparisons (Sec.~\ref{subsec_3d})
\end{abstract}

\section{Implementation Details}
We build WeatherCity upon a dynamic Gaussian scene graph following the node design in OmniRe~\cite{chenomnire}, containing a sky node, a static background node, and multiple rigid and non-rigid object nodes for vehicles and pedestrians, respectively, each represented by 3D Gaussian primitives with learnable position, scale, rotation, opacity, and a shared appearance feature vector. For each weather condition, a lightweight weather-specific MLP with two fully connected layers, ReLU activation, and a final Sigmoid layer maps the shared feature of every Gaussian to its RGB color, yielding a set of weather-dependent Gaussians while keeping the underlying geometry shared across all conditions.
\label{sec_detail}
\subsection{Training Details}
\label{subsec_traing}
\noindent \textbf{Parameter setting.} We optimize all scene nodes jointly for 30,000 iterations using Adam, while adopting node-specific learning rates to stabilize training for different motion patterns.  The rotation parameters of Gaussian nodes are trained with a learning rate of \(5\times10^{-5}\) for non-rigid nodes and \(1\times10^{-5}\) for all other nodes. All other scalar parameters, including shared features and weather-decoder weights, use a base learning rate of \(1\times10^{-4}\).
All Gaussian densification operations are driven by the absolute gradient of the Gaussian parameters~\cite{ye2024absgsrecoveringfinedetails} with a densification threshold of \(3\times10^{-4}\); and the scaling threshold for pruning is set to \(3\times10^{-3}\). The shared Gaussian feature dimension is fixed to 32 with the hidden layer dimension equal to 64. During training, we randomly sample both raw clear-weather images and edited images of different weather types as supervision, and render corresponding views from the Gaussian representation to compute reconstruction losses, while all Gaussians (scene and weather particles) are rasterized with the standard 3D Gaussian Splatting pipeline.

\noindent \textbf{Weather Particle Simulation.} For dynamic weather effects, we instantiate dedicated Gaussian nodes for rain and snow inside a scene-aligned 3D bounding volume, and treat each particle as an elongated or compact Gaussian primitive that is jointly rasterized with the reconstructed scene.  In the rainy setting, we sample 40,000 raindrop particles whose base color is fixed to $c_{\text{rain}} = [0.7, 0.7, 0.8]$, with scale initialized to $[0.0025, 0.0025, 0.075]$ and opacity set to 0.13, which produces thin, semi-transparent streaks aligned with the velocity direction.  For snow, we use 16,000 particles with a brighter color $c_{\text{snow}} = [0.9, 0.9, 0.95]$, an anisotropic scale of $[0.0064, 0.004, 0.004]$, and opacity 0.2, leading to denser and more softly visible flakes that exhibit fluttering motion under the turbulence term.  Fog is modeled as a global depth-dependent medium using the Beer–Lambert formulation, where the transmittance is parameterized by density $d_{f}=0.2$ and the global fog color is set to $c_{\text{fog}} = [0.8, 0.8, 0.85]$, enabling continuous control of visibility and color tone by adjusting $d_{f}$ and $c_{\text{fog}}$.

\noindent \textbf{Prompt design.}
In all experiments, we use identical text instructions for Qwen-Image and all baselines to ensure a fair comparison. As shown in Table \ref{tab:weather_prompts}, for each target weather condition, we design a structured prompt that separately specifies: (1) strict preservation of the original layout and style, (2) the desired visual properties of the target weather, and (3) prohibited artifacts.

The prompts enforce consistent content preservation—including camera composition, object categories, and spatial arrangement—so the models focus solely on modifying global atmospheric conditions.

For rain, the prompt specifies an overcast sky, wet roads with puddles and reflections, and cool, dim lighting, while forbidding sunlight, dry ground, and hallucinated objects.
For snow, it additionally removes all but the designated white and red vehicles, converts foliage to snow-covered bare branches, and requires overcast lighting with falling snow, without introducing new elements or distortions.
For fog, it similarly keeps only the white and red vehicles and requests realistic atmospheric haze with depth-dependent visibility reduction and soft, overcast illumination, while prohibiting clear-air or warm-light appearances. These prompts ensure consistent, content-preserving weather editing across rain, snow, and fog for all compared methods.

\begin{table*}[t]
    \centering
    \caption{Prompts used for Qwen-Image and all baseline methods under each weather condition.}
    \label{tab:weather_prompts}\renewcommand{\arraystretch}{1.25}
    \begin{tabularx}{1.0\textwidth}{c |X}
        \hline
        Weather & \multicolumn{1}{c}{Prompt} \\
        \hline
        \multirow{6}{*}{Rainy } 
         &
        Please strictly maintain the original composition, all scene contents (including ground, buildings, vegetation, cars, background, pedestrians, etc.), their positions, and the original artistic style. Convert the scene to a rainy setting. Requirements: The image should be clear and realistic; the sky must be overcast with dark clouds; the ground should be wet with puddles and reflections; the lighting should be dim, and the overall tone should be cool to create a rainy atmosphere. Do NOT include: sunny weather, blue skies and white clouds, sunlight, dry ground, any elements not present in the original image, trees that are not in the original, distorted visuals, deformed subjects, or incorrect proportions. \\
        [0.3em]
        \hline
        \multirow{8}{*}{Snowy} &
        Please strictly maintain the original composition, all scene contents (including ground, buildings, vegetation, cars, background, pedestrians, etc.), their positions, and the original artistic style. Convert the scene to a snowy setting. Requirements: The image should be clear and realistic; the sky should be overcast with falling snowflakes; the ground should be naturally covered with snow; do not add any extra vegetation; the lighting should be soft, and the overall tone should be cool. If the original image contains green leaves, please turn them into snow-covered bare branches. Do NOT include: sunny weather, warm tones, sunlight, melting snow, elements unrelated to the original image, elements not present in the original image, distorted visuals, deformed subjects, or incorrect proportions. \\[0.3em]
        \hline
    \multirow{7}{*}{Foggy} &
        Please strictly maintain the original composition, all scene contents (including ground, buildings, vegetation, cars, background, pedestrians, etc.), their positions, and the original artistic style. Convert the scene to a foggy setting. Requirements: The image should be clear while maintaining realistic atmospheric fog; the sky should appear overcast; the scene should be filled with natural, soft fog that reduces visibility in the distance; lighting should be diffused and soft, with an overall cool tone. Do NOT include: sunny weather, warm tones, sunlight, dry and clear air, elements unrelated to the original image, elements not present in the original image, deformed subjects, or incorrect proportions. \\
        [0.3em]
        \hline
        \multirow{7}{*}{Snowy 
        \& vehicle removal} & 
        Please remove all vehicles in the image except for the white and red ones, and then transform the scene into snowy weather. Requirements: The image should be clear and realistic; the sky should be overcast with falling snowflakes; the ground should be naturally covered with snow; do not add any extra vegetation; the lighting should be soft, and the overall tone should be cool. If the original image contains green leaves, please turn them into snow-covered bare branches. Do NOT include: sunny weather, warm tones, sunlight, melting snow, elements unrelated to the original image, distorted visuals, deformed subjects, or incorrect proportions.\\
        \hline  
    \end{tabularx}
    
\end{table*}

\subsection{Loss Functions}
\label{subsec_loss_functions}
To jointly optimize all learnable parameters of the scene representation and the dynamic nodes model, we employ a weighted combination of image-based reconstruction terms and regularization losses, 
\begin{equation}
\begin{aligned}
\mathcal{L}_{total} = & \mathcal{L}_{rgb} + \lambda_{cc}\mathcal{L}_{cc} + \lambda_{depth}\mathcal{L}_{depth} \\
 & + \lambda_{opacity}\mathcal{L}_{opacity}+\mathcal{L}_{reg}.
\end{aligned}
\end{equation}
We set the depth weight to $\lambda_{\text{depth}} = 0.01$, the opacity weight to $\lambda_{\text{opacity}} = 0.05$, and the content consistency weight to $\lambda_{\text{cc}} = 1.0$. The losses $\mathcal{L}_{rgb}$, $\mathcal{L}_{cc}$, and $\mathcal{L}_{depth}$ have been introduced in the main text. Here, we additionally present the details of losses $\mathcal{L}_{opacity}$ and $\mathcal{L}_{reg}$.
\paragraph{Opacity loss.} We further constrain the opacities of the Gaussians using a 2D supervision derived from the sky mask. 
For each view, we render an opacity map $O_{G}$ from the current Gaussian scene, and use a binary sky mask $M_{\text{sky}}$ obtained from semantic segmentation~\cite{xie2021segformer}. The opacity loss takes the form
\begin{equation}
\begin{aligned}
\mathcal{L}_{\text{opacity}}
=
&-\sum_{u} O_{G}(u) \log O_{G}(u) \\
&-\sum_{u} M_{\text{sky}}(u) \log\bigl(1 - O_{G}(u)\bigr).
\end{aligned}
\end{equation}

\paragraph{Regularization loss.} 
The regularization loss comprises sharp shape regularization, voxel deformer regularization, temporal smoothness regularization, and scaling regularization.

\subsection{Baselines}
\label{subsec_base}
\noindent \textbf{ControlNet~\cite{zhang2023adding}} is an image editing method that introduces spatial conditional control into text-to-image diffusion models. Its core lies in achieving precise guidance of the image editing process by injecting spatial constraint information, supporting image modification tasks under various conditions. By aligning the intermediate features of pre-trained diffusion models with spatial conditions (such as edges and depth), this method maintains the flexibility of text prompts while enhancing the structural consistency of editing results. In the weather editing task of this study, ControlNet ~\cite{zhang2023adding} generates target weather effects based on text prompts. However, experimental results indicate that it is prone to scene content distortion (e.g., vehicle deformation, incorrect lane markings), lacks fine-grained control over weather intensity, and has a slow inference speed (only 0.033 FPS), making it difficult to meet the real-time and consistency requirements of 4D scene simulation.

\noindent \textbf{TurboEdit~\cite{deutch2024turboedit}} is a text-guided image editing method based on few-step diffusion models. It aims to reduce artifacts generated during diffusion model editing and improve editing efficiency and image quality by optimizing noise scheduling strategies and novel guidance techniques. By adjusting the noise distribution and guidance signals during the diffusion process, this method maintains the visual coherence of editing results while reducing inference steps, making it suitable for fast image content modification tasks. As an image-level editing comparison baseline, TurboEdit~\cite{deutch2024turboedit} can generate weather effects to a certain extent. However, limited by the nature of 2D image editing, it cannot model depth-aware atmospheric effects (e.g., depth attenuation of fog), and exhibits insufficient performance in semantic consistency and scene structure preservation. Meanwhile, its inference speed is still far below real-time requirements (0.097 FPS).

\noindent \textbf{FRESCO~\cite{yang2024fresco}} is a video editing model for zero-shot video translation. Its core innovation lies in modeling spatial-temporal correspondence to achieve cross-domain editing of video sequences without specialized training for specific tasks. By capturing spatial alignment and temporal coherence between video frames, this method completes style or scene transformation of video content under the guidance of text prompts, suitable for dynamic sequence editing tasks. In this study, FRESCO~\cite{yang2024fresco} serves as a video-level editing baseline to verify the weather transformation capability in dynamic scenes. However, experimental results show that it still suffers from significant scene content distortion in multi-weather editing, and has limited ability to simulate dynamic weather effects (e.g., falling rain and snow). Its inference speed (0.142 FPS) is difficult to support the real-time simulation needs.

\noindent \textbf{Qwen-Image~\cite{wu2025qwen}} is a powerful text-guided image editing foundation model with high-quality image generation and editing capabilities. It can accurately respond to semantic requirements in text prompts and generate realistic and content-consistent editing results. Trained on large-scale data, this model achieves a good balance between image content preservation and target effect generation, supporting flexible editing in various scenarios. 
However, as a pure 2D image editing model, Qwen-Image~\cite{wu2025qwen} lacks modeling of temporal coherence between frames. When used alone, it is prone to temporal flickering and geometric inconsistency issues, and cannot support object-level editing and dynamic weather simulation of 4D scenes.

\noindent \textbf{ClimateNeRF~\cite{li2023climatenerf}} is a 3D-level weather editing method that integrates physical simulation with Neural Radiance Fields (NeRF) to enable the editing of various climate effects in 3D scenes.  
By leveraging the inherent 3D geometric modeling capability of NeRF, this method achieves more realistic environmental rendering compared to 2D image editing approaches.  
However, a key limitation is that it is confined to static reconstruction and simulation of static weather phenomena, it lacks the ability to model dynamic vehicles and cannot simulate dynamic weather effects, such as falling rain or snow, which are critical for 4D urban scene simulation. Additionally, it supports only a limited range of weather editing operations and fails to realize flexible text-guided weather control. Furthermore, ClimateNeRF~\cite{li2023climatenerf} has a rendering speed of 0.032 FPS, which is insufficient to meet the demands of real-time simulation.

\begin{figure*}[t] 
\center
\includegraphics[width=1.0\textwidth]{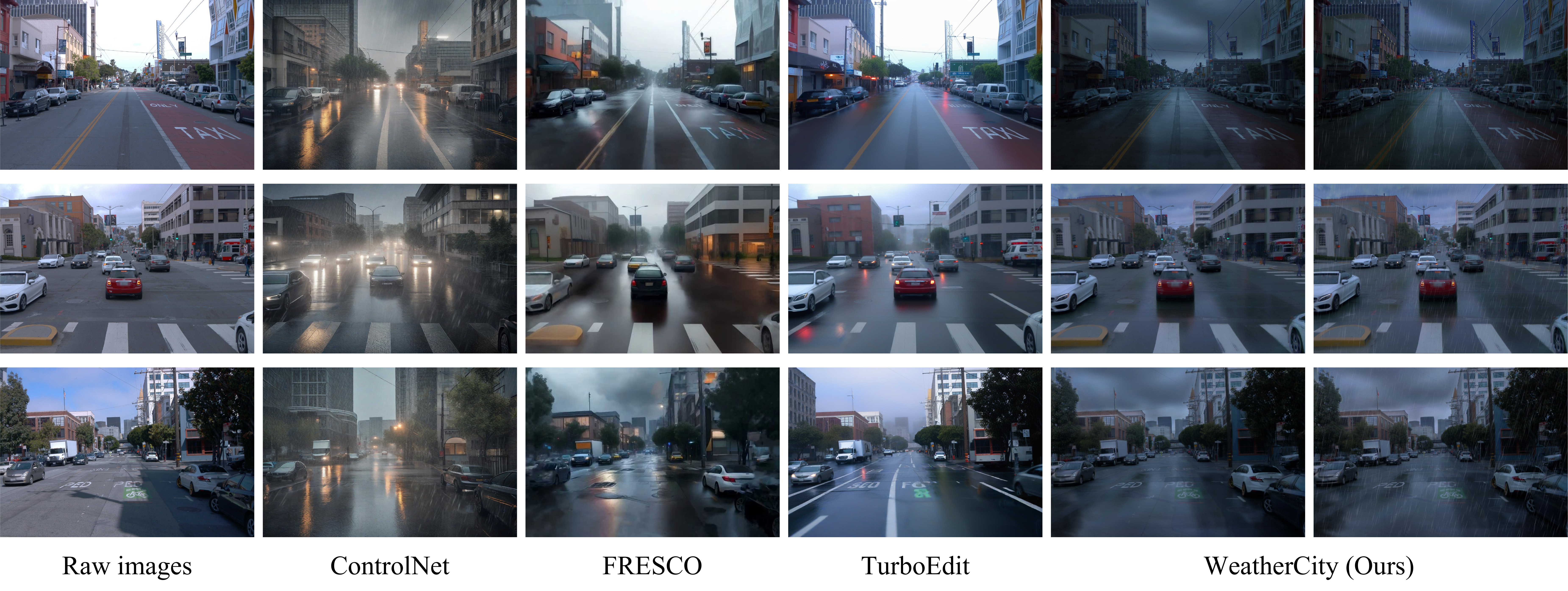}
\caption{\textbf{Qualitative comparison of rainy weather on the Waymo Open Dataset.} Our model excels at capturing complex lighting interactions and environmental changes induced by rain, providing a higher level of physical realism.}
\label{rainy_waymo}
\end{figure*}

\begin{figure*}[t] 
\center
\includegraphics[width=1.0\textwidth]{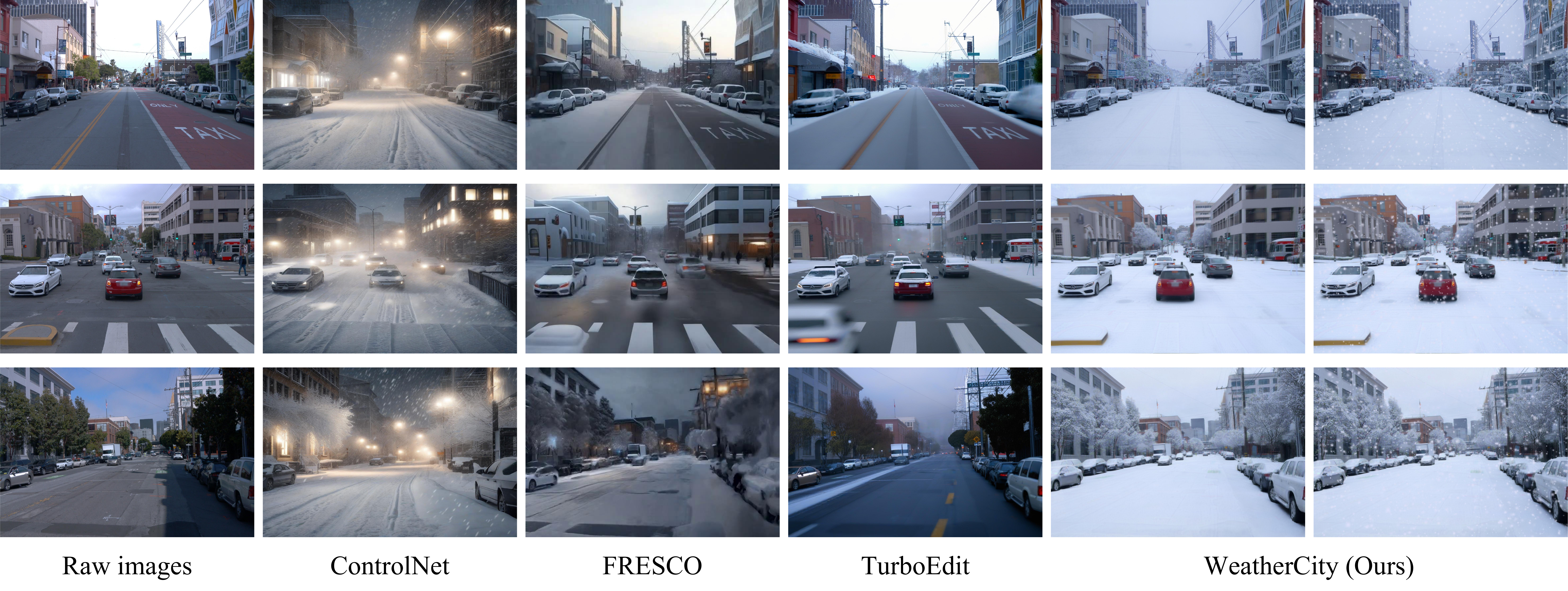}
\caption{\textbf{Qualitative comparison of snowy weather on the Waymo Open Dataset.} As shown, WeatherCity generates highly realistic snow accumulation on vehicles and roads while avoiding the texture distortions common in other approaches.}
\label{snowy_waymo}
\end{figure*}

\begin{figure*}[t] 
\center
\includegraphics[width=1.0\textwidth]{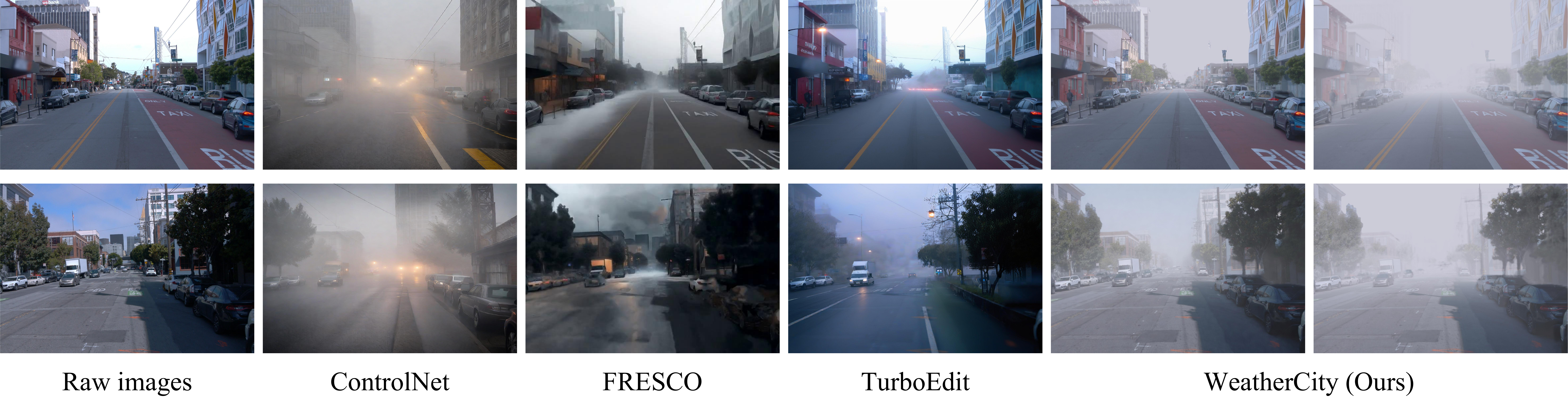}
\caption{\textbf{Qualitative comparison of foggy weather on the Waymo Open Dataset.} WeatherCity demonstrates superior depth-consistent haze rendering, effectively preserving the semantic layout of the original scene compared to baseline methods.}
\label{foggy_waymo}
\end{figure*}

\begin{figure*}[t] 
\center
\includegraphics[width=1.0\textwidth]{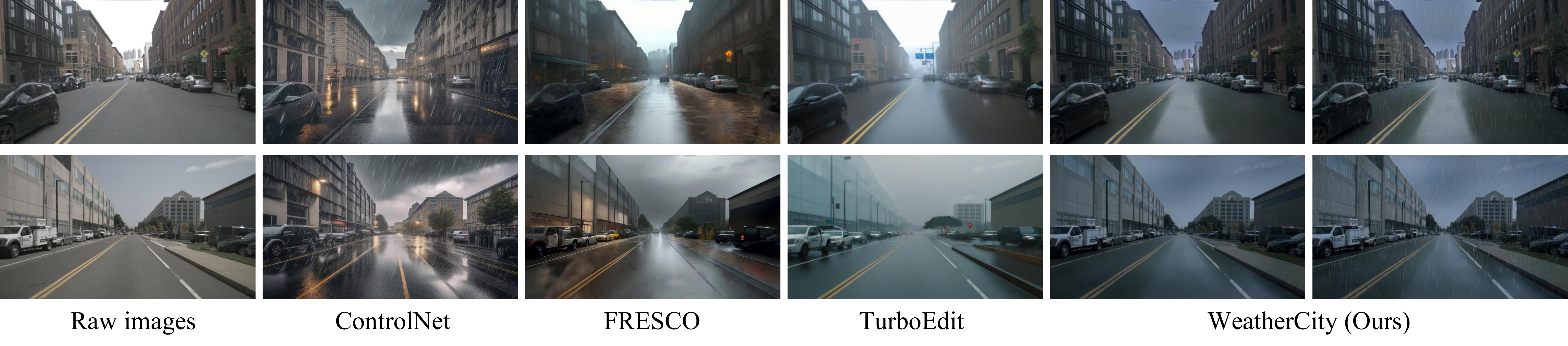}
\caption{\textbf{Qualitative comparison of rainy weather on the nuScenes Dataset.} Our method accurately simulates realistic wet surface reflections and rain streaks, significantly outperforming ControlNet and FRESCO in terms of visual fidelity.}
\label{rainy_nuscenes}
\end{figure*}

\begin{figure*}[h] 
\center
\includegraphics[width=1.0\textwidth]{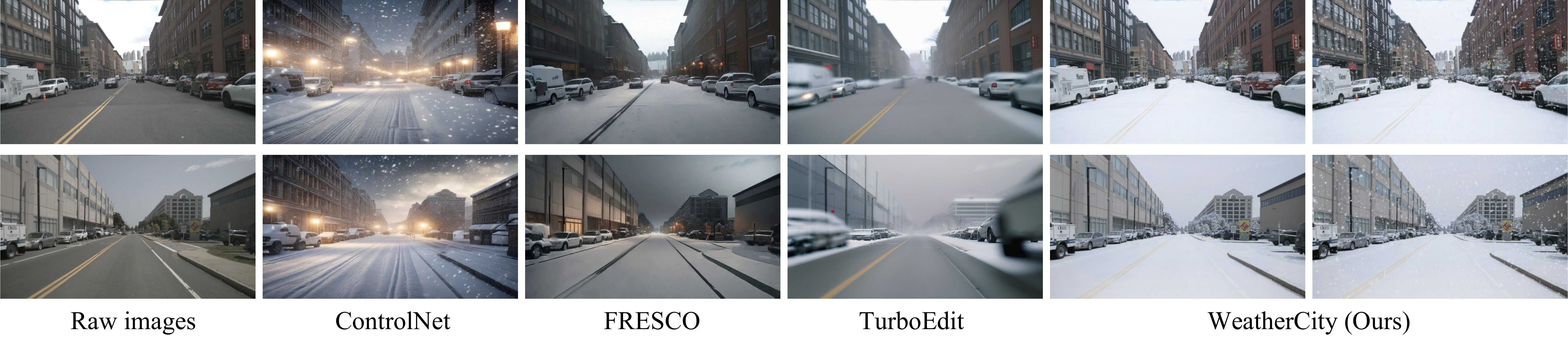}
\caption{\textbf{Qualitative comparison of snowy weather on the nuScenes Dataset.} Our approach produces the most visually plausible winter scenes with natural white-out effects, surpassing the consistency and quality of competing editing frameworks.}
\label{snowy_nuscenes}
\end{figure*}

\begin{figure*}[h] 
\center
\includegraphics[width=1.0\textwidth]{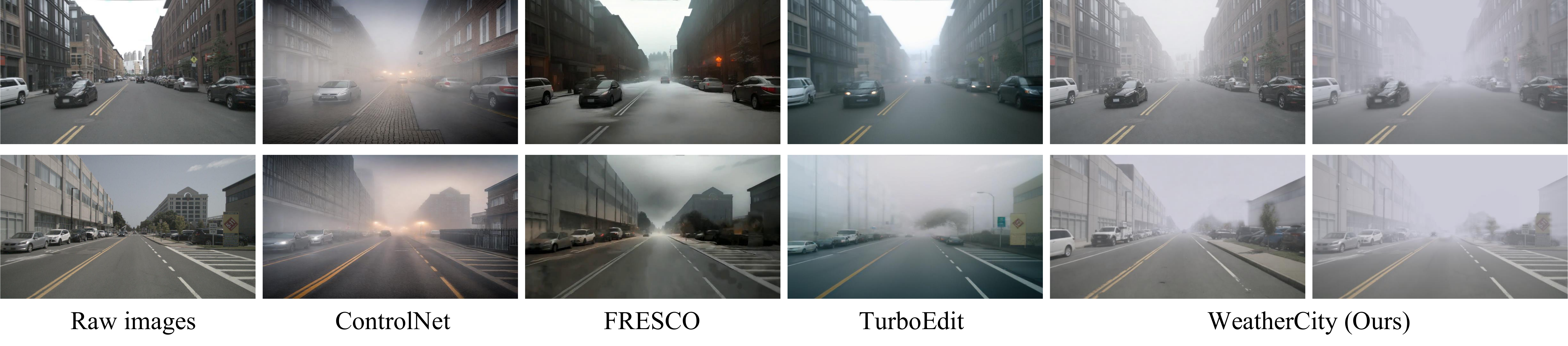}
\caption{\textbf{Qualitative comparison of foggy weather on the nuScenes Dataset.} WeatherCity achieves a natural atmospheric haze that smoothly obscures distant objects, exhibiting fewer artifacts than TurboEdit or other baselines.}
\label{foggy_nuscenes}
\end{figure*}

\begin{figure*}[t] 
\center
\includegraphics[width=1.0\textwidth]{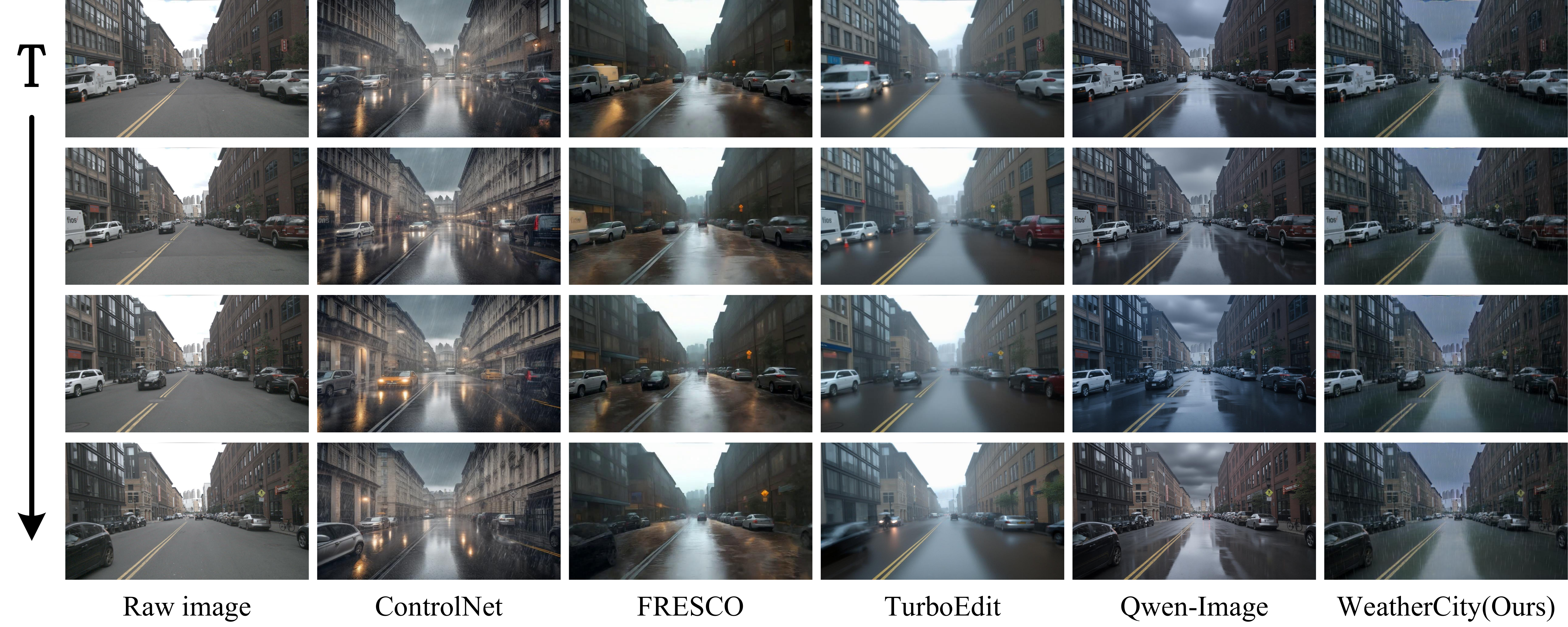}
\caption{\text{Qualitative comparison of rainy weather on temporal consistency.} The baselines suffer from scene deformation and erratic fluctuations in weather effects, exhibiting noticeable inter-frame flickering. In contrast, our approach enables temporally consistent weather editing.}
\label{rainy_time_series}
\end{figure*}

\begin{figure*}[t] 
\center
\includegraphics[width=1.0\textwidth]{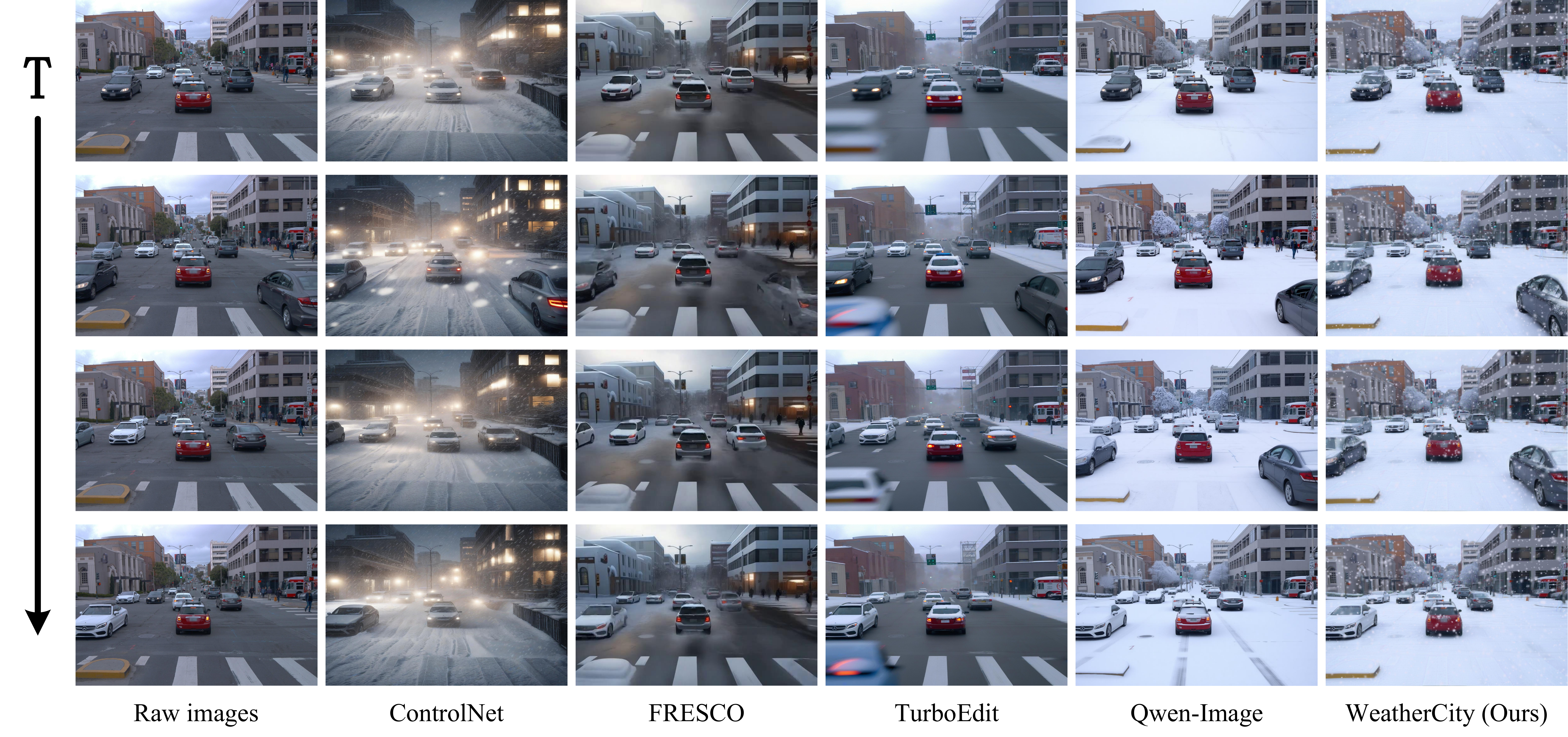}
\caption{\text{Qualitative comparison of snowy weather on temporal consistency.} The baselines suffer from scene deformation and erratic fluctuations in weather effects, exhibiting noticeable inter-frame flickering. In contrast, our approach enables temporally consistent weather editing.}
\label{snowy_time_series}
\end{figure*}

\begin{figure*}[t] 
\center
\includegraphics[width=1.0\textwidth]{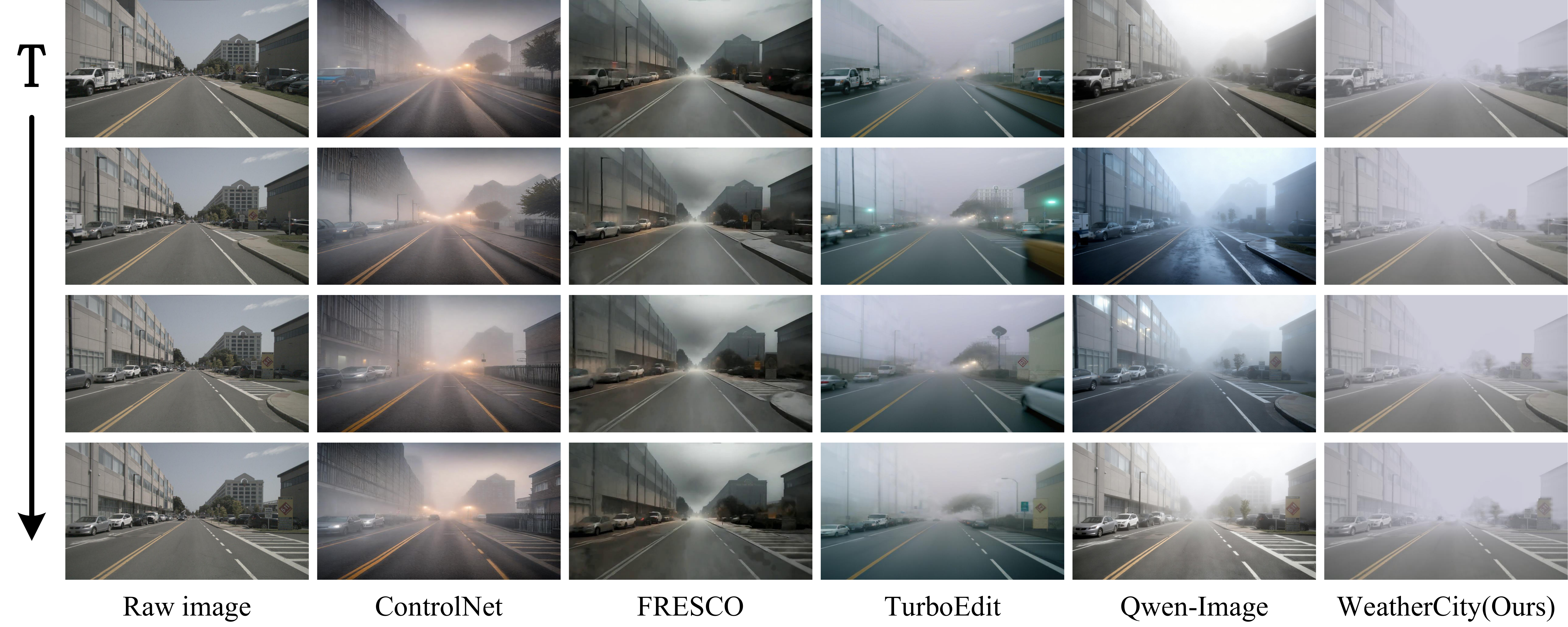}
\caption{\text{Qualitative comparison of foggy weather on temporal consistency.} The baselines suffer from scene deformation and erratic fluctuations in weather effects, exhibiting noticeable inter-frame flickering. In contrast, our approach enables temporally consistent weather editing.}
\label{foggy_time_series}
\end{figure*}

\subsection{Evaluation Details}
\label{subsec_eval}
\noindent \textbf{CLIP-Score (CLIP-S~\cite{hessel2021clipscore}).}
CLIP-S measures the visual similarity between the original image $I$ and the edited image $\hat{I}$ using the CLIP image encoder.
Let $f_{\text{CLIP}}(\cdot)$ denote the CLIP model, then the metric is computed as:
\begin{equation}
    \text{CLIP-S} = 
\frac{
\langle f_{\text{CLIP}}(I),\, f_{\text{CLIP}}(\hat{I}) \rangle
}{
\| f_{\text{CLIP}}(I) \|_2 \, \| f_{\text{CLIP}}(\hat{I}) \|_2
}.
\end{equation}

\noindent \textbf{CLIP Directional Similarity (CLIP-DS~\cite{gal2022stylegan}).}
CLIP-DS evaluates whether the “editing direction’’ in CLIP space—produced by the edited image relative to the original image—aligns with the target editing direction defined by the text prompt.
Given the original image $I$, edited image $\hat{I}$, and target text instruction $T$, the metric is:
\begin{equation}
    \text{CLIP-DS}
=
\frac{
\left\langle 
f_{\text{CLIP}}(\hat{I}) - f_{\text{CLIP}}(I),\;
f_{\text{CLIP}}(T)
\right\rangle
}{
\| f_{\text{CLIP}}(\hat{I}) - f_{\text{CLIP}}(I) \|_2 \,
\| f_{\text{CLIP}}(T) \|_2 }.
\end{equation}

\noindent \textbf{Semantic Consistency Score (Sem-CS).}
Sem-CS measures the semantic consistency between edited and original images. 
We apply a \texttt{ConvNeXt-XL-384$\times$384}~\cite{liu2022convnet} model pretrained on ADE20K~\cite{Zhou_2017_CVPR} to perform panoptic segmentation on the original image $I$ and the edited image $\hat{I}$. 
Let $\text{IoU}_c$ denote the IoU of category $c$, aggregated over all ADE20K classes $\mathcal{C}$. 
Sem-CS is defined as the frequency-weighted IoU (fwIoU):
\begin{equation}
    \text{Sem-CS} = 
\frac{
\sum_{c \in \mathcal{C}} 
n_c \, \text{IoU}_c
}{
\sum_{c \in \mathcal{C}} n_c
},
\end{equation}
where $n_c$ is the number of pixels belonging to class $c$ in the ground-truth segmentation of the original image.

We note that fog synthesis substantially reduces scene visibility, which consequently invalidates metrics designed for content preservation (e.g., CLIP-S, Sem-CS). Thus, for foggy weather, our evaluation is solely based on the CLIP-DS metric.

\section{Additional Results and Analysis}
\label{sec_res}
\subsection{Detailed Quantitative and Qualitative Results}
Table~\ref{table:waymo_results} and Table~\ref{table:nuscenes_results} present the complete quantitative comparison results for the Waymo and nuScenes datasets, respectively. Our method significantly outperforms all baseline approaches (ControlNet, FRESCO, and TurboEdit) across all metrics. Specifically, higher CLIP-S indicates better content preservation w.r.t. the original scene. Furthermore, the improvements in Sem-CS quantify our method's ability to preserve the original scene content—such as road layout and vehicle geometry—during the weather transformation process, confirming that WeatherCity minimizes the content distortion often observed in image-level editing frameworks.

\noindent \textbf{Rainy:} As illustrated in Fig.~\ref{rainy_waymo} and Fig.~\ref{rainy_nuscenes}, our method successfully renders high-frequency details such as falling raindrops and specular reflections on wet road surfaces. Unlike baselines such as ControlNet and FRESCO, which tend to apply a global style transfer that often blurs the boundary between the road and the environment, our method leverages 3D scene representations to ensure that reflections are geometrically consistent with the camera view.
    
\noindent \textbf{Snowy:} In snowy weather generation, as shown in Fig.~\ref{snowy_waymo} and Fig.~\ref{snowy_nuscenes}, our method achieves realistic snow accumulation on distinct surfaces, such as vehicle roofs and vegetation, without altering the underlying object semantics. The visual evidence shows that competitive methods (e.g., ControlNet) frequently hallucinate structures or warp the shape of vehicles when attempting to add snow textures. WeatherCity effectively avoids these artifacts, preserving the clear contours of dynamic objects and lane markings.

\noindent \textbf{Foggy:} The foggy scenarios highlight the advantage of our depth-aware approach. As seen in Fig.~\ref{foggy_waymo} and Fig.~\ref{foggy_nuscenes}, WeatherCity simulates physically plausible depth attenuation, where visibility decreases naturally with distance. In contrast, baselines like TurboEdit and FRESCO often apply a uniform haze layer or introduce artifacts that obscure nearby objects, failing to respect the scene's depth map.
    
Overall, while baseline methods can produce general weather-like appearances, they suffer from severe content distortion—manifesting as warped vehicles and erroneous scene structures. WeatherCity overcomes these limitations, offering a robust solution for high-fidelity, geometry-preserving weather simulation.

\label{subsec_res}

\begin{table*}[t]
\caption{Comparison on Waymo Open Dataset. $\uparrow$ means higher is better.}
\label{table:waymo_results}
\centering
\renewcommand{\arraystretch}{1.15}
\begin{tabular}{lccccccc}
\toprule
\multirow{2}{*}{\textbf{Method}} & \multicolumn{3}{c}{\textbf{Rainy}} &
\multicolumn{3}{c}{\textbf{Snowy}} & \multicolumn{1}{c}{\textbf{Foggy}} \\
\cmidrule(lr){2-4}\cmidrule(lr){5-7}\cmidrule(lr){8-8}
& CLIP-S$\uparrow$ & CLIP-DS$\uparrow$ & Sem-CS$\uparrow$ 
& CLIP-S$\uparrow$ & CLIP-DS$\uparrow$ & Sem-CS$\uparrow$ 
& CLIP-DS$\uparrow$ \\

\midrule

ControlNet~\cite{zhang2023adding} 
& 0.654 & 0.228 & 0.713 & 0.615 & 0.261 & 0.677 & 0.225 \\

TurboEdit~\cite{deutch2024turboedit} 
& 0.843 & 0.233 & 0.825 & 0.816 & 0.228 & 0.787 & 0.221 \\

FRESCO~\cite{yang2024fresco} 
& 0.721 & 0.209 & 0.852 & 0.719 & 0.253 & 0.797 & 0.177 \\

Qwen-Image~\cite{wu2025qwen} 
& 0.813 & 0.248 & 0.845 & 0.757 & 0.310 & 0.840 & 0.251 \\

\textbf{WeatherCity (Ours)} 
& \textbf{0.898} & \textbf{0.300} & \textbf{0.931} 
& \textbf{0.847} & \textbf{0.330} & \textbf{0.899} & \textbf{0.278} \\

\bottomrule
\end{tabular}
\end{table*}

\begin{table*}[t]
\caption{Comparison on nuScenes Dataset. $\uparrow$ means higher is better.}
\label{table:nuscenes_results}
\centering
\begin{tabular}{lccccccc}
\toprule
\multirow{2}{*}{\textbf{Method}} & \multicolumn{3}{c}{\textbf{Rainy}} &
\multicolumn{3}{c}{\textbf{Snowy}} & \multicolumn{1}{c}{\textbf{Foggy}} \\
& CLIP-S$\uparrow$ &CLIP-DS$\uparrow$ & Sem-CS$\uparrow$ & CLIP-S$\uparrow$ &CLIP-DS$\uparrow$ & Sem-CS$\uparrow$ & CLIP-DS$\uparrow$   \\
\midrule

ControlNet~\cite{zhang2023adding} 
& 0.703 & 0.250 & 0.810 & 0.609 & 0.234 & 0.812 & 0.201 \\

TurboEdit~\cite{deutch2024turboedit} 
& 0.806 & 0.225 & 0.848 & 0.758 & 0.261 & 0.811 & 0.266 \\

FRESCO~\cite{yang2024fresco}
& 0.726 & 0.220 & 0.863 & 0.694 & 0.239 & 0.847 & 0.213 \\

Qwen-Image~\cite{wu2025qwen} 
& 0.823 & 0.256 & 0.891 & 0.785 & 0.302 & 0.913 & 0.264 \\

\textbf{WeatherCity (Ours)} 
&\textbf{0.880} & \textbf{0.272} &\textbf{0.977} & \textbf{0.860} & \textbf{0.333} & \textbf{0.960} & \textbf{0.301}  \\

\bottomrule
\end{tabular}

\end{table*}

\subsection{Temporal Consistency Comparison}
\label{subsec_tem}
Temporal consistency is crucial for 4D urban scene simulation, requiring coherent motion of dynamic objects and continuous evolution of weather effects across frames. Qualitative comparisons on Waymo/nuScenes dynamic sequences (Fig.~\ref{rainy_time_series}, Fig.~\ref{snowy_time_series}, Fig.~\ref{foggy_time_series}) reveal that baseline methods generally suffer from inter-frame inconsistency: scene geometry undergoes deformation and dynamic vehicles exhibit shape changes between consecutive frames, while weather effects display random fluctuations with noticeable inter-frame flickering. In contrast, our approach achieves temporally consistent weather editing throughout the temporal sequence.


\subsection{3D Baseline Comparison}
\label{subsec_3d}
\begin{figure*}[t] 
\center
\includegraphics[width=1.0\linewidth]{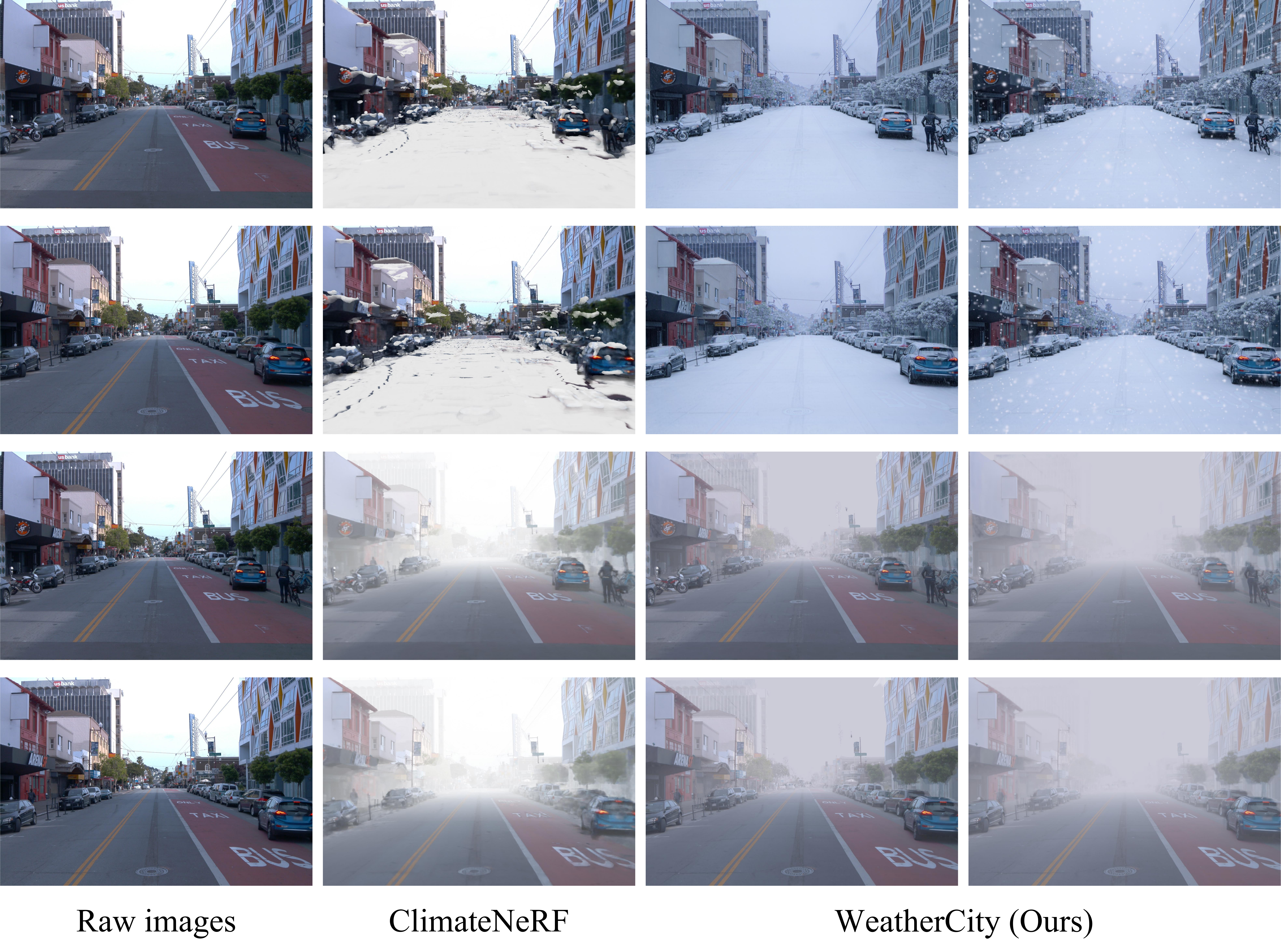}
\caption{Qualitative comparison with ClimateNeRF on Waymo Open Dataset. ClimateNeRF is limited to static modeling and static weather effect editing, while WeatherCity produces more photorealistic results and additionally supports dynamic weather particles, such as falling snowflakes.}
\label{fig_climatenerf}
\end{figure*}

We compare WeatherCity with ClimateNeRF~\cite{li2023climatenerf}, a NeRF-based 3D weather editing method, with quantitative and qualitative results presented in Tab.~\ref{tab_climatenerf} and Fig.~\ref{fig_climatenerf}. ClimateNeRF exhibits significant limitations due to its static 3D representation: it cannot effectively model dynamic objects (resulting in motion-blurred vehicles) nor simulate evolving weather effects. In contrast, WeatherCity, leveraging dynamic Gaussian modeling, outperforms ClimateNeRF across all metrics while delivering more realistic editing effects. Our approach additionally supports dynamic weather particles (such as falling snowflakes) and achieves significantly higher rendering efficiency than ClimateNeRF, conclusively validating its superiority.

\begin{table*}[t]
    \caption{Comparison on Waymo Open Dataset scene 788. $\uparrow$ means higher is better.}
    \label{tab_climatenerf}
    \centering
    \renewcommand{\arraystretch}{1.15}
    \begin{tabular}{lccccc}
    \toprule
    \multirow{2}{*}{\textbf{Method}} & 
    \multicolumn{3}{c}{\textbf{Snowy}} & \multicolumn{1}{c}{\textbf{Foggy}} & \multirow{2}{*}{\textbf{FPS}$\uparrow$}\\
    \cmidrule(lr){2-4}\cmidrule(lr){5-5}
    
    & CLIP-S$\uparrow$ & CLIP-DS$\uparrow$ & Sem-CS$\uparrow$ 
    & CLIP-DS$\uparrow$ \\
    
    \midrule
    
    ClimateNeRF~\cite{li2023climatenerf} 
    & 0.807& 0.294 & 0.905 & 0.269 & 0.032 \\
    
    \textbf{WeatherCity (Ours)} 
    & \textbf{0.847} & \textbf{0.341} & \textbf{0.941} & \textbf{0.280} & \textbf{25.67}\\
    
    \bottomrule
    \end{tabular}
\end{table*}